\pdfoutput=1
\documentclass{article}
\usepackage[T1]{fontenc}
\usepackage{iclr2027_conference,times}

\usepackage{amsmath,amsfonts,bm}

\def\eqref#1{equation~\ref{#1}}

\def\1{\bm{1}}

\DeclareMathAlphabet{\mathsfit}{\encodingdefault}{\sfdefault}{m}{sl}
\SetMathAlphabet{\mathsfit}{bold}{\encodingdefault}{\sfdefault}{bx}{n}

\usepackage{hyperref}
\usepackage{url}
\usepackage{graphicx}
\usepackage{booktabs}
\usepackage{array}
\usepackage{xcolor}
\usepackage{colortbl}
\usepackage{caption}
\usepackage{listings}
\usepackage{needspace}
\newcolumntype{C}[1]{>{\centering\arraybackslash}p{#1}}
\newcolumntype{P}[1]{>{\raggedleft\arraybackslash}p{#1}@{\,/\,}>{\raggedright\arraybackslash}p{#1}}
\hypersetup{
  colorlinks=true,
  linkcolor=[rgb]{0.12,0.29,0.56},
  citecolor=[rgb]{0.12,0.29,0.56},
  urlcolor=[rgb]{0.12,0.29,0.56},
  pdftitle={rTaste: Do Research Agents Have Taste? Evaluating Early Stage Research Judgment},
  pdfauthor={Art Kanke},
}

\title{rTaste: Do Research Agents Have Taste?\\ Evaluating Early Stage Research Judgment}

\author{{Art Kanke} \\
{University of Minnesota} \\
\texttt{kanke011@umn.edu}
}

\iclrfinalcopy
\begin{document}

\maketitle
\lhead{Preprint}

\begin{abstract}

Do research agents have taste, the judgment of which experiments are worth running before their outcomes are known? We give six frontier models seven recent NLP papers. Each model, working as a research agent alone or as a team, proposes minimum viable experiments (MVEs), forecasts whether each will pass a threshold it sets and ranks them. A separate executor then attempts every MVE, including those ranked last, and three model auditors judge the results. Of 415 MVEs, 194 are judged sound, valuable and new, but fewer than half of these (73) are also judged beyond the seed paper and non obvious. The agents' first choices match the auditors' most valued experiment more often than a random pick, but are not shown to improve on choosing the experiment written first. Their forecasts beat a base rate but are not measurably more accurate than those of a fresh instance of the same model given only the paper and registrations. Teams cost more without a clear overall improvement in proposal quality. The agents produce many worthwhile experiments, but the evidence that their rankings improve the choice of which to run remains limited. The evaluation harness and a dataset of all 415 MVEs, with their forecasts and audited outcomes, are available at \url{https://github.com/ArtKanke/rTaste}.

\end{abstract}

\section{Introduction}\label{sec:intro}

Large language model agents can now carry out much of the empirical research workflow, from literature search and implementation to running experiments and writing reports \citep{lu2024aiscientist, schmidgall2025agentlab}. Frontier AI labs report that agents write much of their code and support a rapidly growing number of experiments \citep{favaro2026aibuildsitself, openai2026acceleration}. A risk assessment of agents inside four frontier labs, however, found no evidence that any of them relied on agents to set research agendas or make budget decisions \citep{metr2026frontierrisk}. The step that has not been delegated is close to what researchers call taste, the judgment of which questions and experiments are worth pursuing before their outcomes are known \citep{hamming1986research}. In practice this judgment is often exercised through small experiments. A researcher tests a direction cheaply and lets the result decide whether a larger study is needed. Such an experiment is useful only if its outcome is not obvious in advance and the researcher's expectation about it is accurate enough to act on. An agent that proposes experiments with predictable outcomes or that is confidently wrong about what its experiments will show spends a limited budget poorly even if it runs every experiment correctly.

Existing evaluations address parts of this question. Execution benchmarks specify a task and measure how well an agent completes it \citep{chan2025mlebench, starace2025paperbench}. Studies of research ideas often judge proposals before execution, although those judgments can change when the ideas are carried out \citep{si2026ideationexecution}. Other work tests whether models can predict outcomes or choose among experiments on a given problem \citep{wen2025predicting, foster2026rpm, zhang2026datasense}. We examine these judgments for experiments that agents propose from a recent paper and compare their expectations with the outcomes of every proposal.

\begin{figure}[t]
\centering
\includegraphics[width=\textwidth]{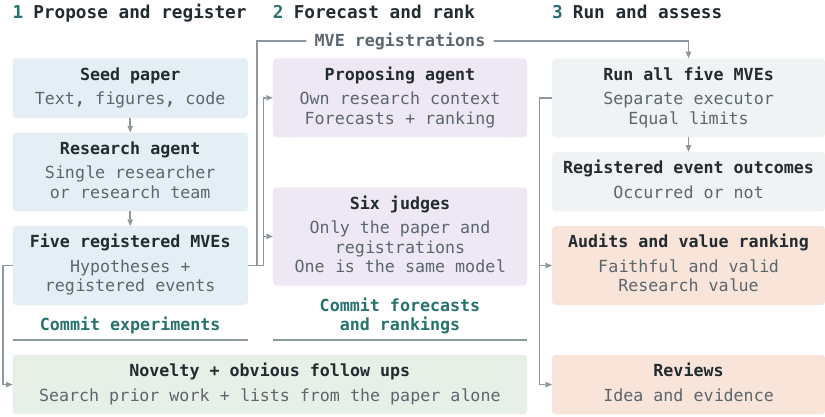}
\caption{The study design.}
\label{fig:design}
\end{figure}

We ask what experiments agents propose from a new paper, how accurately they predict the outcomes, and how well they choose among their proposals. We give each model a recent NLP paper, the \emph{seed paper}, with its code. Working as a research agent, once as a single researcher and once as a research team, the model proposes five \emph{minimum viable experiments} (MVEs), inexpensive tests meant to indicate whether a direction deserves further investment. Before any experiment runs, the agent forecasts whether each MVE will meet its registered threshold and ranks the five (Figure~\ref{fig:design}). A separate executor then attempts every MVE, including those the agent ranked last. Three language models audit the results. We also compare the agent with a fresh instance of the same model that receives only the seed paper and registrations and makes the same forecasts and ranking.

Across seven seed papers and 415 MVEs, 194 are judged sound, valuable and new. On average the four larger models propose about three such MVEs in five, and the two smaller models, GPT-6 Luna and Muse Spark 1.3, about one. Two models, GPT-6 Astra and GPT-6.1 Sol, propose most of the MVEs judged beyond the seed paper and non obvious. A research team does not, on average, propose measurably better MVEs than a single researcher of the same model. The agents judge their own proposals less well. Their forecasts are more accurate than a base rate, but their own reading and search did not measurably improve them over those of the same model judge. The agent's first choice is the auditors' most valued MVE more often than a random pick would be. We find no evidence that it is better than the MVE the agent wrote first or than the same model judge's first choice. We release the evaluation harness and the dataset of all 415 MVEs at \href{https://github.com/ArtKanke/rTaste}{github.com/ArtKanke/rTaste}.

\section{Related Work}\label{sec:related}

\paragraph{Research agents and benchmarks.} Research agent systems now span the workflow from ideation to written reports \citep{yamada2025aiscientistv2, miyai2026jraiscientist}, some organized around a persistent tree of hypotheses and isolated executors \citep{jin2026arbor}. RE-Bench, AIRS-Bench, and DeepRSI give agents a task with a fixed metric and score the result \citep{wijk2025rebench, lupidi2026airsbench, deeprsi2026}. TasteVal, concurrent with our work, measures how efficiently a model's own experiments reach an expert's score on a fixed task \citep{jaffe2026tasteval}. EXP-Bench asks agents to reproduce experiments from published papers \citep{kon2025expbench}. Other benchmarks give agents research problems drawn from recent or unpublished work \citep{garikaparthi2026researchgym, kim2026intotheunknown}. In each, the problem is given to the agent, whereas our agents choose what to test from a seed paper. Studies of such agents report frequent failures to complete experiments, reported results that do not match the underlying code \citep{chen2025mlrbench, zhang2026trueautoresearch}, and weak judgment about when evidence is sufficient \citep{kirgis2026crux}.

\paragraph{Evaluating research ideas.} A second group of work evaluates research ideas directly. Blind comparisons with expert researchers found that large language model ideas were rated more novel than human ideas \citep{si2025novelideas}, and studies of expert preferences found that automated judges reproduce those preferences only partially \citep{chen2026ideationarena} or little better than chance \citep{baig2026taste}. Judgments of this kind are made without outcomes. When ideas from the earlier study were executed by researchers, the advantage of the model generated ideas disappeared \citep{si2026ideationexecution}. Language models asked to judge the soundness of proposals are also strongly biased toward approval \citep{ho2026soundnessbench}. Agent generated ideas tend to stay close to their source literature \citep{tang2026narrow}. Other studies judge research foresight against the later literature, asking which directions will be pursued or will gain attention \citep{li2026ideaforecast, tian2026foresci, ye2026proofoftime}. Another study compares a model's proposed next step in a project with the step its authors actually took \citep{gong2026researchtrails}.

\paragraph{Predicting experimental outcomes.} The work closest to ours predicts experimental outcomes before execution. Models can predict which of two published ideas performs better on a benchmark \citep{mule2026forecast}, which of two candidate programs will score higher \citep{zheng2026predict}, or the numeric result of an experiment from its description \citep{park2026anticipatory}. Recent work also asks models which training recipe will reach the lowest loss \citep{ren2026architectureiq} and how much a change to a training or inference recipe will improve a benchmark score \citep{wang2026rwm}. Some register expected outcomes before pilot experiments and run only the best supported hypotheses in full \citep{ren2026evigraph}, so their expectations about the other hypotheses are never checked at full scale. Others run every candidate, to train a model that scores hypotheses \citep{vasu2025harpa} or to learn from large batches of executed ideas \citep{si2026executiongrounded}. Where calibration has been measured for agents, on coding tasks with clear success criteria, agents are overconfident about their own success \citep{kaddour2026agenticuncertainty, barkan2025capable}.

\paragraph{Research teams.} Several research systems divide the work among specialized agents that generate, critique, and rank hypotheses \citep{gottweis2026coscientist, schmidgall2026gemini}, and a lead agent coordinating parallel subagents has improved performance on research style information tasks at a substantially higher token cost \citep{hadfield2025multiagentresearch}. Studies suggest that such gains depend on the task and on how coordination is organized \citep{kim2025scalingagents} and that when total computation is held equal a single agent often matches or exceeds multi agent designs \citep{tran2026singleagent}.

\section{Experimental Setup}\label{sec:setup}

\subsection{Seed papers}\label{sec:seeds}

We use seven NLP papers (arXiv cs.CL) with released code, taken from the daily papers list on Hugging Face\footnote{\url{https://huggingface.co/papers}} and first made public between 3 August and 24 September 2026. We chose papers this recent so that little follow up work exists for an agent to find and repeat. The selection also excludes work that requires large GPU training runs or models with more than 8B parameters. The full list can be found in Appendix~\ref{app:setup}. Agents receive the paper's full text, an image of every page, and read access to its code repository. We did not provide a summary, a suggested research gap or direction, or a hypothesis.

\subsection{Agents and harnesses}\label{sec:policies}

We evaluate six models at maximum reasoning effort: GPT-6 Astra, Claude Fable 5.1, Claude Opus 5.5, GPT-6.1 Sol, GPT-6 Luna, and Muse Spark 1.3.\footnote{A reply cut off at the output limit is requested once more at medium effort. This happened for the lead of the Claude Opus 5.5 team on six of the seven seed papers, so 30 of the 415 registrations were written at medium effort.} Evaluation was conducted between 6 and 8 October 2026. Each model works under two harnesses, so each seed paper has twelve \emph{conditions}. We refer to GPT-6 Luna and Muse Spark 1.3 as the two smaller models and to the other four as the larger models. The judges (Section~\ref{sec:protocol}) also run at maximum effort, and the executors, novelty judges, auditors, and reviewers at high effort.

A \emph{single researcher} (SR) works in one continuous context from reading the seed paper to committing its forecasts. A \emph{research team} (RT) divides the same work among separate instances of the same model. Two proposers each read the seed paper, search the literature, and draft experiments independently. A critic examines the pooled drafts once for scientific validity and for feasibility. A lead then keeps, revises, or merges the drafts into five experiments, commits them, and makes the team's forecasts. The critic and the lead have no tools, so the lead has not itself searched the literature or read the code. The comparison asks whether two drafts and one round of critique help a model propose better MVEs than one continuous context. Under both harnesses, agents have the same limits and search the literature with the same arXiv tool. A comparison between models is therefore not also a comparison between their providers' web tools.

\subsection{Protocol}\label{sec:protocol}

The agent makes two commitments before any experiment runs (Figure~\ref{fig:design}). First, it proposes five minimum viable experiments. Each MVE is specified in a short registration with a hypothesis, its closest prior work, and one \emph{registered event}: a stated statistic passes a threshold set in advance. Second, the agent gives each MVE a probability that its registered event will occur and ranks the five from most to least worth pursuing. We call the MVE it ranks highest its \emph{first choice}. The proposing agent knows that all proposed MVEs will run, so its ranking states a priority rather than deciding what runs.

We also ask each of the six models, as a \emph{judge} in a new context, for the same forecasts and ranking. They see the seed paper and the five registrations. We do not provide the code, the agent's reasoning or forecasts, or which model proposed the MVEs, and they cannot search. We call the judge from the agent's own model its \emph{same model judge}. We run the MVEs of a seed paper only after all twelve conditions have committed.

After the experiments are committed and before execution, two \emph{novelty judges}, Claude Opus 5.5 and GPT-6.1 Sol, use the same literature search tool to label each MVE as equivalent to existing work, a close variant, or distinct. An MVE labeled equivalent or a close variant still runs. To see whether an MVE is an obvious follow up, we also ask each of the six models, given only the seed paper and no search, to list ten follow up experiments it would propose. The two novelty judges check each MVE against these lists, and we count it as an \emph{obvious follow up} if either finds it there.

\subsection{Execution and evaluation}\label{sec:execution}

\paragraph{Executors.} One of two executors, Claude Opus 5.5 or GPT-6.1 Sol, implements each MVE in a new context. The same two executors serve all six models, so a smaller model's proposals are not held back by its own implementation. The executor sees the seed paper, its code, and the one MVE it implements. We did not provide the other four MVEs, the ranking, the forecasts, or the name of the model that proposed the MVE. All MVEs of one model on a seed paper go to the same executor and the executors alternate across seed papers. Because both executors are among the six models, 58 of the 412 delivered MVEs were implemented by a separate instance of the model that proposed them. We run every MVE with the same limits, no matter its rank. The executor works on a separate machine with one A100 GPU for at most 45 minutes, may run open models of up to 8B parameters, and may not call hosted models.\footnote{On the seventh seed paper, 53 of its 60 MVEs were implemented and rerun on H200 machines with the same limits and prompts.} If it refuses or exceeds its limits, the other executor makes one independent attempt. If both fail, the MVE is not delivered.

\paragraph{Evaluation.} We then delete the results that came with the executor's submission and rerun its evaluation on a new machine the executor never used. The rerun determines whether the registered event occurred. An MVE is \emph{delivered} when the rerun ends without error and writes a number. The rerun repeats the executor's own code, so whether that code is right is left to the auditors.

\paragraph{Audits.} Three auditors, Claude Opus 5.5, GPT-6.1 Sol, and Grok 4.7, judge every delivered MVE. Grok 4.7 is not one of the six models we study and does not execute. The auditors see the registration, the executor's notes, the output of the rerun, and the files it left. They do not see the seed paper or who proposed or implemented the MVE. Each auditor judges whether the MVE is faithful to its registration and valid and rates its research value from 0 to 3. We take the majority of the three judgments of faithfulness and validity and the median of the three research values. We call an MVE \emph{sound} when it is faithful and valid, \emph{valuable} when its research value is 2 or 3 and \emph{new} when neither novelty judge found close prior work. The auditors also rank the five MVEs of each condition by research value. The MVE with the best mean rank is the auditors' \emph{most valued MVE}, against which we score the agent's first choice.

\paragraph{Reviews.} After the study, we added a review of every delivered MVE. Three reviewers, the same models as the auditors, read the seed paper's text, the registration, and what the auditors were shown, but none of the forecasts, rankings, audits, or novelty labels. They answered fixed questions: how the MVE relates to the seed paper, how obvious its idea is, and what the evidence says about the hypothesis.

\section{Findings}\label{sec:findings}

Of the 84 conditions, 83 produced MVEs.\footnote{On one seed paper, the provider of Claude Opus 5.5 refused the requests of its single researcher before the agent proposed anything.} We make each comparison within seed papers and report its average over the seven seed papers, with a 95\% confidence interval (CI) computed across them.

\subsection{Quality of the MVEs}\label{sec:quality}

\begin{table}[t]
\centering
\caption{Model comparison, single researcher (SR) and research team (RT). Sound, valuable and new: mean number of the five MVEs of a condition. Beyond the paper and non obvious, rated non obvious, and on a list of obvious follow ups: numbers of MVEs. Brackets: 95\% CI over seed papers. $^{\dagger}$Its single researcher ran on six seed papers (30 MVEs); in the first column, RT minus SR counts the seventh as zero.}
\label{tab:models}
\resizebox{\textwidth}{!}{%
\begin{tabular}{@{}l*{2}{C{18pt}}*{2}{C{14.5pt}}*{2}{C{12.5pt}}*{2}{C{16pt}}*{2}{C{18pt}}*{2}{C{17.5pt}}*{2}{C{18pt}}*{2}{C{12.5pt}}@{}}
\toprule
 & \multicolumn{2}{c}{Sound,} & \multicolumn{2}{c}{Beyond} & \multicolumn{2}{c}{} & \multicolumn{2}{c}{On a} & \multicolumn{2}{c}{} & \multicolumn{2}{c}{Registered} & \multicolumn{2}{c}{} & \multicolumn{2}{c@{}}{Sound,} \\
 & \multicolumn{2}{c}{valuable} & \multicolumn{2}{c}{the paper,} & \multicolumn{2}{c}{Rated} & \multicolumn{2}{c}{list of} & \multicolumn{2}{c}{Informative} & \multicolumn{2}{c}{event} & \multicolumn{2}{c}{Cost per} & \multicolumn{2}{c@{}}{valuable} \\
 & \multicolumn{2}{c}{and new,} & \multicolumn{2}{c}{non} & \multicolumn{2}{c}{non} & \multicolumn{2}{c}{obvious} & \multicolumn{2}{c}{evidence,} & \multicolumn{2}{c}{occurred,} & \multicolumn{2}{c}{condition,} & \multicolumn{2}{c@{}}{and new} \\
 & \multicolumn{2}{c}{of five} & \multicolumn{2}{c}{obvious} & \multicolumn{2}{c}{obvious} & \multicolumn{2}{c}{follow ups} & \multicolumn{2}{c}{\% of sound} & \multicolumn{2}{c}{\% of sound} & \multicolumn{2}{c}{dollars} & \multicolumn{2}{c@{}}{per \$100} \\
\cmidrule(lr){2-3}\cmidrule(lr){4-5}\cmidrule(lr){6-7}\cmidrule(lr){8-9}\cmidrule(lr){10-11}\cmidrule(lr){12-13}\cmidrule(lr){14-15}\cmidrule(l){16-17}
 & SR & RT & SR & RT & SR & RT & SR & RT & SR & RT & SR & RT & SR & RT & SR & RT \\
\midrule
GPT-6 Astra & 2.57 & 3.29 & 21 & 21 & 25 & 24 & 9 & 7 & 55 & 55 & 24 & 16 & 16.7 & 30.0 & 15 & 11 \\
Claude Fable 5.1 & 2.43 & 2.86 & 3 & 4 & 4 & 9 & 20 & 17 & 67 & 80 & 57 & 53 & 25.9 & 46.1 & 9 & 6 \\
Claude Opus 5.5$^{\dagger}$ & 3.33 & 3.14 & 2 & 8 & 7 & 17 & 18 & 20 & 87 & 72 & 53 & 44 & 24.0 & 37.4 & 14 & 8 \\
GPT-6.1 Sol & 3.00 & 3.71 & 20 & 22 & 22 & 24 & 9 & 10 & 61 & 88 & 27 & 33 & 10.0 & 13.2 & 30 & 28 \\
GPT-6 Luna & 0.71 & 2.00 & 6 & 7 & 6 & 8 & 21 & 17 & 33 & 44 & 33 & 25 & 9.1 & 8.1 & 8 & 25 \\
Muse Spark 1.3 & 0.14 & 1.00 & 5 & 7 & 6 & 8 & 11 & 22 & 39 & 45 & 39 & 38 & 7.4 & 8.9 & 2 & 11 \\
\midrule
All & 2.00 & 2.67 & 57 & 69 & 70 & 90 & 88 & 93 & 57 & 64 & 38 & 35 & 15.3 & 23.9 & 13 & 11 \\
RT minus SR & \multicolumn{2}{c}{$+0.71$} & \multicolumn{2}{c}{$+12$} & \multicolumn{2}{c}{$+20$} & \multicolumn{2}{c}{$+5$} & \multicolumn{2}{c}{$+7.0$} & \multicolumn{2}{c}{$-3.7$} & \multicolumn{2}{c}{$+8.6$} & \multicolumn{2}{c@{}}{$-2$} \\
 & \multicolumn{2}{c}{\makebox[0pt]{[$-0.10$, $1.52$]}} & & & & & & & & & & & & & & \\
\bottomrule
\end{tabular}}
\end{table}

We count an MVE as sound, valuable and new when it meets all three (Section~\ref{sec:execution}). Nearly all MVEs were delivered and most were judged sound, so most of those that fall short repeat prior work or have low research value. Because the seed papers are recent, new mostly means not yet published: most MVEs on a list of obvious follow ups are also new. The four larger models propose about three such MVEs in five and the two smaller models about one. The gap holds for single researchers and for teams, while the order among the four larger models changes from one seed paper to another.

The smaller models' MVEs often run as registered but leave the question open. For each sound MVE, the reviewers stated what the evidence says about the hypothesis, and we call the evidence \emph{informative} when at least two of them read it as for, against, or mixed. MVEs of GPT-6 Luna are sound as often as those of GPT-6.1 Sol, but their evidence is informative about half as often (38\% against 74\%), and that of Muse Spark 1.3 is similar (Table~\ref{tab:models}).

\subsection{Kinds of experiments}\label{sec:kinds}

The agent is asked to use the seed paper as inspiration for novel research and it chooses the question itself. To see which questions it chooses, we use the reviewers' answers on how each MVE relates to the seed paper (Table~\ref{tab:kinds}) and the lists of obvious follow ups that each model wrote from the seed paper alone (Section~\ref{sec:protocol}).

\begin{table}[t]
\centering
\caption{The kind of experiment each model proposes, SR / RT: counts of delivered MVEs by the answer that at least two of the three reviewers give.}
\label{tab:kinds}
\small
\setlength{\tabcolsep}{4pt}
\begin{tabular}{@{}lP{14.5pt}P{23.5pt}P{14.5pt}P{14.5pt}P{26pt}@{}}
\toprule
 & \multicolumn{2}{c}{Checks} & \multicolumn{2}{c}{Reruns, varies} & \multicolumn{2}{c}{Builds} & \multicolumn{2}{c}{New} & \multicolumn{2}{c@{}}{No two} \\
MVEs of & \multicolumn{2}{c}{the paper} & \multicolumn{2}{c}{or explains} & \multicolumn{2}{c}{on} & \multicolumn{2}{c}{direction} & \multicolumn{2}{c@{}}{reviewers agree} \\
\midrule
GPT-6 Astra & 7 & 7 & 0 & 0 & 24 & 23 & 4 & 4 & 0 & 0 \\
Claude Fable 5.1 & 8 & 13 & 2 & 0 & 24 & 21 & 0 & 0 & 1 & 1 \\
Claude Opus 5.5 & 16 & 17 & 0 & 2 & 13 & 16 & 0 & 0 & 1 & 0 \\
GPT-6.1 Sol & 4 & 4 & 1 & 0 & 26 & 22 & 3 & 6 & 1 & 2 \\
GPT-6 Luna & 4 & 5 & 1 & 1 & 29 & 28 & 0 & 1 & 1 & 0 \\
Muse Spark 1.3 & 6 & 8 & 5 & 2 & 24 & 21 & 0 & 1 & 0 & 2 \\
\midrule
All & 45 & 54 & 9 & 5 & 140 & 131 & 7 & 12 & 4 & 5 \\
\bottomrule
\end{tabular}
\end{table}

Most MVEs build on the seed paper rather than check it, and few open a new direction. By the reviewers' answers, 66\% of the delivered MVEs build on the paper, 24\% check one of its claims, and 5\% open a direction that the paper merely suggests (Table~\ref{tab:kinds}). Building on the paper is often the obvious next step: 44\% of all MVEs match an experiment on one of the lists of obvious follow ups. The agents also converge on the same experiments. On one seed paper \citep{nguyentri2026flashdllm}, a method for faster decoding whose appendix gives a correctness argument, all eight conditions of the four larger models wrote the same check of that argument first.

Two models more often go beyond the paper with an experiment that is not obvious. We call an MVE \emph{non obvious} when at least two reviewers say that few or none of ten researchers would list it and that the paper neither reports nor names it, and \emph{beyond the paper and non obvious} when at least two reviewers each say this and also that it builds on the paper or opens a new direction. This rating comes from the reviewers and differs from the lists of obvious follow ups: a fifth of these MVEs are also on a list. GPT-6 Astra and GPT-6.1 Sol proposed two thirds of them (Table~\ref{tab:models}), and they lead for each reviewer separately. Within seed papers, their share of such MVEs exceeds that of Claude Fable 5.1 and Claude Opus 5.5 by 48 percentage points (95\% CI 29 to 66) and is higher on all seven seed papers. GPT-6 Luna, a smaller model from the same provider, builds on the paper as often, but the reviewers rate few of its ideas non obvious. MVEs beyond the paper and non obvious are sound, valuable and new at least as often as the others, but their registered events occur less often. Of the 194 sound, valuable and new MVEs, 73 are beyond the paper and non obvious. The reviewers recommend a larger study more often for the MVEs of Claude Fable 5.1 and Claude Opus 5.5 than for those of GPT-6 Astra and GPT-6.1 Sol.

\subsection{Registered events and evidence}\label{sec:events}

Each registration commits the agent to one registered event before the MVE runs and it is natural to read the event as the success of the experiment. For each sound MVE, we compare whether the event occurred with the reviewers' reading of the evidence.

Whether the event occurs depends on how the agent defines it. In most of their MVEs, GPT-6 Astra and GPT-6.1 Sol compare a bound of a confidence interval with the threshold, rather than a point estimate (Table~\ref{tab:app-event-forms}), and the other four models rarely do. The events of GPT-6 Astra and GPT-6.1 Sol occur in 25\% of their sound MVEs, compared with 52\% for Claude Fable 5.1 and Claude Opus 5.5 (Table~\ref{tab:models}). On one seed paper \citep{liang2026auroralm}, four conditions tested whether a model's own prediction errors harm decoding more than random errors of the same size, and all three reviewers read all four results as support for it. Three of the conditions had set the threshold at a difference of 5 percentage points and their events occurred. The fourth had set it at 8 points, and its event did not.

When the registered event does not occur, the evidence is often still informative. The reviewers judged the evidence informative for 50\% of the sound MVEs whose event did not occur, mostly as evidence against the hypothesis. In the other direction, they judged the evidence inconclusive for 20\% of the sound MVEs whose event occurred.

\subsection{Forecasts}\label{sec:forecasts}

We next ask whether the agent that designed an MVE forecasts its registered event better than an outside reader. The outside reader is the same model judge, which sees only the seed paper and the five registrations. The agent's own reading and search did not measurably improve its forecasts over this reader's: the difference between their Brier scores (lower is better) is $-$0.006, with a 95\% CI from $-$0.021 to 0.009 (Table~\ref{tab:brier}).

\begin{table}[t]
\centering
\caption{Brier scores by whose MVEs are forecast (rows) and who forecasts (columns), lower is better. Judges: A: GPT-6 Astra, F: Claude Fable 5.1, O: Claude Opus 5.5, S: GPT-6.1 Sol, L: GPT-6 Luna, M: Muse Spark 1.3. Underlined: the same model judge. Avg.: the Brier score of the mean of the six judges' probabilities.}
\label{tab:brier}
\small
\setlength{\tabcolsep}{3.5pt}
\begin{tabular}{@{}lccccccccc}
\toprule
MVEs & Event &  & \multicolumn{7}{c}{Judge} \\
\cmidrule(lr){4-10}
of & rate & Agent & A & F & O & S & L & M & Avg. \\
\midrule
GPT-6 Astra & .20 & \cellcolor{gray!1}.120 & \cellcolor{gray!0}\underline{.117} & \cellcolor{gray!6}.136 & \cellcolor{gray!5}.134 & \cellcolor{gray!3}.128 & \cellcolor{gray!23}.193 & \cellcolor{gray!20}.185 & \cellcolor{gray!5}.133 \\
Claude Fable 5.1 & .55 & \cellcolor{gray!31}.222 & \cellcolor{gray!21}.187 & \cellcolor{gray!37}\underline{.241} & \cellcolor{gray!31}.222 & \cellcolor{gray!21}.186 & \cellcolor{gray!33}.229 & \cellcolor{gray!37}.241 & \cellcolor{gray!26}.205 \\
Claude Opus 5.5 & .48 & \cellcolor{gray!24}.199 & \cellcolor{gray!28}.211 & \cellcolor{gray!32}.223 & \cellcolor{gray!26}\underline{.203} & \cellcolor{gray!26}.204 & \cellcolor{gray!34}.231 & \cellcolor{gray!33}.228 & \cellcolor{gray!25}.202 \\
GPT-6.1 Sol & .30 & \cellcolor{gray!16}.169 & \cellcolor{gray!14}.163 & \cellcolor{gray!23}.193 & \cellcolor{gray!17}.173 & \cellcolor{gray!14}\underline{.164} & \cellcolor{gray!21}.187 & \cellcolor{gray!12}.157 & \cellcolor{gray!13}.160 \\
GPT-6 Luna & .29 & \cellcolor{gray!40}.251 & \cellcolor{gray!20}.184 & \cellcolor{gray!18}.178 & \cellcolor{gray!16}.170 & \cellcolor{gray!26}.203 & \cellcolor{gray!39}\underline{.247} & \cellcolor{gray!25}.202 & \cellcolor{gray!21}.187 \\
Muse Spark 1.3 & .38 & \cellcolor{gray!22}.189 & \cellcolor{gray!27}.208 & \cellcolor{gray!20}.184 & \cellcolor{gray!20}.183 & \cellcolor{gray!28}.209 & \cellcolor{gray!26}.203 & \cellcolor{gray!29}\underline{.214} & \cellcolor{gray!21}.188 \\
\midrule
All & .37 & \cellcolor{gray!22}.192 & \cellcolor{gray!18}.179 & \cellcolor{gray!22}.192 & \cellcolor{gray!19}.181 & \cellcolor{gray!20}.182 & \cellcolor{gray!29}.216 & \cellcolor{gray!26}.205 & \cellcolor{gray!19}.179 \\
\bottomrule
\end{tabular}
\end{table}

Both forecasts are more accurate than a base rate that gives every MVE the event rate of the other six seed papers (Brier score 0.240), and the mean of the six judges' probabilities is more accurate than the agent (0.179 against 0.192). The agent has seen more than its registrations state, but it is not shown to know more about its events than its registrations tell a reader. The agents are also not uniformly overconfident: GPT-6 Astra, GPT-6.1 Sol, GPT-6 Luna and Muse Spark 1.3 state higher probabilities than the rates at which their events occur, while Claude Fable 5.1 and Claude Opus 5.5 state lower ones.

\subsection{First choices}\label{sec:first}

The agent's first choice is the auditors' most valued MVE in 38\% of conditions, where a random pick gives 20\%. The MVE that the agent wrote first and the same model judge's first choice are the most valued about as often, and we find no evidence that the agent's first choice beats either: its advantage over the MVE written first is 3 percentage points (95\% CI $-$8 to 15). The first choice is also sound, valuable and new no more often than a random pick.

Agents register their obvious experiments first: the MVE written first is on a list of obvious follow ups in 61\% of conditions, against 35\% for the fifth. The agents then rank largely in the order in which they wrote them, and their first choice is the MVE they wrote first in 47\% of conditions, where 20\% would be expected if order did not matter. The event of the first choice occurs in 47\% of conditions and that of the last choice in 29\%, so the first choice is often an experiment likely to come out as registered. The auditors also rank sound MVEs whose event occurred higher, mostly because these MVEs give clearer evidence.

The judges rank in the same way: they too put the MVE written first on top in 39\% of conditions on average. They and the auditors also rank obvious follow ups higher, and an agent's ranking agrees more with the same model judge and with other judges from the same provider than with judges from other providers. On average, a judge's first choice is an MVE beyond the paper and non obvious less often than a random pick would be, while the agent's own first choice is one about as often.

\subsection{Single researchers and teams}\label{sec:teams}

A team proposes more sound, valuable and new MVEs than a single researcher of the same model, 0.71 more of five on average, but the difference is not statistically significant (95\% CI $-$0.10 to 1.52; Table~\ref{tab:models}). The difference is largest for the two smaller models, about one more of five, at about the cost of a single researcher. Claude Opus 5.5 working alone does as well as its team.

A team changes little in what is proposed. Its MVEs check the paper, match the lists of obvious follow ups, and are beyond the paper and non obvious about as often as a single researcher's. They are rated non obvious somewhat more often (Table~\ref{tab:models}), but the 95\% CI includes zero. What the critic and the lead change is the design of each experiment: teams more often compare a confidence bound with the threshold and state lower probabilities.

\subsection{Execution and resources}\label{sec:resources}

\begin{table}[t]
\centering
\caption{Execution tokens and compute by model, SR / RT.}
\label{tab:compute}
\small
\setlength{\tabcolsep}{4pt}
\begin{tabular}{@{}lP{24pt}P{23.5pt}P{26pt}P{16pt}@{}}
\toprule
 & \multicolumn{4}{c}{Executor} & \multicolumn{4}{c@{}}{GPU minutes} \\
\cmidrule(lr){2-5}\cmidrule(l){6-9}
 & \multicolumn{2}{c}{Output tokens,} & \multicolumn{2}{c}{Working time,} & \multicolumn{2}{c}{} & \multicolumn{2}{c@{}}{} \\
MVEs of & \multicolumn{2}{c}{thousands} & \multicolumn{2}{c}{minutes} & \multicolumn{2}{c}{Implementation} & \multicolumn{2}{c@{}}{Rerun} \\
\midrule
GPT-6 Astra & 18.5 & 23.9 & 13.4 & 13.2 & 9.7 & 9.5 & 5.3 & 4.9 \\
Claude Fable 5.1 & 18.2 & 24.5 & 23.9 & 23.2 & 21.4 & 21.2 & 14.8 & 13.2 \\
Claude Opus 5.5 & 21.0 & 26.2 & 22.8 & 27.9 & 21.1 & 25.7 & 15.1 & 17.0 \\
GPT-6.1 Sol & 16.5 & 22.2 & 12.1 & 11.8 & 9.2 & 7.7 & 5.3 & 4.2 \\
GPT-6 Luna & 13.0 & 13.9 & 14.2 & 10.3 & 12.1 & 7.5 & 5.3 & 4.1 \\
Muse Spark 1.3 & 8.9 & 11.7 & 9.8 & 12.5 & 7.4 & 9.9 & 5.0 & 5.7 \\
\midrule
All & 15.9 & 20.4 & 15.9 & 16.5 & 13.3 & 13.6 & 8.3 & 8.2 \\
\bottomrule
\end{tabular}
\end{table}

A condition cost \$15.31 on average for a single researcher and \$23.93 for a team, counting the agent's model calls and the implementation and rerun of its five MVEs (Table~\ref{tab:models}). Per \$100, single researchers produced 13 sound, valuable and new MVEs and teams 11, and GPT-6.1 Sol produced the most. The cost of proposing differs about a hundredfold between models, from GPT-6 Luna to Claude Fable 5.1, while the cost of implementing and rerunning an MVE differs about twofold, since the same two executors served every model. Without the requirement of novelty, a team of GPT-6 Luna produced as many sound and valuable MVEs per \$100 as GPT-6.1 Sol, but about half as many beyond the paper and non obvious, where GPT-6.1 Sol produced the most.

The executors delivered 397 of the 415 MVEs on their first attempt, and the limits were rarely reached: the median implementation took 12 minutes, well within its limits (Appendix~\ref{app:limits}). The agents reached their tool call limit in 5\% of conditions. The MVEs of Claude Fable 5.1 and Claude Opus 5.5 take more than twice as much GPU time to implement as those of GPT-6 Astra and GPT-6.1 Sol (Table~\ref{tab:compute}). The auditors judge 18\% of the MVEs implemented by the Claude Opus 5.5 executor not sound, compared with 3\% of those implemented by the GPT-6.1 Sol executor, while the evidence is informative equally often under both.

\section{Discussion}\label{sec:discussion}

We began this study to see how many good small experiments frontier models can propose when a recent paper is their source of inspiration. By a good MVE we mean one that is small and not obvious and that opens a direction worth a larger study. Most of the experiments were sound, but most either tested the seed paper itself or extended it in a way that many researchers would propose, and few opened a new direction. Current agents appear better suited to examining a paper and taking its obvious next steps than to moving beyond it. That is useful in itself, since a new paper's claims can be checked under controls and stricter measurements for a few dollars each, but it is a different capability from proposing new research, and evaluations of research agents should measure the two separately.

The models differ less in how often their experiments are sound than in what kind of experiment they propose. GPT-6 Astra and GPT-6.1 Sol more often go beyond the paper with an experiment that is not obvious and register stricter tests, which pass less often. Claude Fable 5.1 and Claude Opus 5.5 stay closer to the paper and register tests that pass more often. The experiments of the two smaller models more often leave the question open. Choosing a model is therefore also choosing a kind of experiment. Because the price of proposing differs far more between models than the cost of running an MVE, a small model working as a team with a capable executor yields about as many sound and valuable experiments per dollar as the most efficient large model, though fewer beyond the paper.

In our study every experiment ran with the same limits, so the agents' rankings stated priorities and did not decide what would be implemented. Those rankings followed the order in which the agents wrote their MVEs and favored the ones likely to pass. A natural next study gives an agent a fixed research budget to allocate among its own MVEs and to revise after the first results, and asks whether the budget goes to the MVEs that turn out sound, valuable and non obvious. The same budget given to human researchers would show how their allocations compare and provide a human baseline. This would extend the evaluation from what an agent proposes and expects to how it spends a limited budget, the decision in which researchers exercise their taste.

\section{Limitations}\label{sec:limitations}

All labels in this study come from models, so we lack a human baseline. The auditors judged soundness and value after seeing the outcomes and two of the three are models we also evaluate and use as executors. Every MVE was attempted regardless of its rank, so the ranking did not decide what ran. An agent whose ranking decides what runs may rank with more reasoning. Each registration had to be complete enough for a separate executor, so part of the forecasting result may be built into the design. Each condition ran once, so we have no measure of how much a result would change on a second run.

\section{Conclusion}\label{sec:conclusion}

We studied whether research agents can evaluate and predict the outcomes of their own experiments before they are run. Given a recent NLP paper, six frontier models proposed small experiments, forecast whether the result of each would pass a threshold they set and ranked them. A separate executor attempted every proposed experiment.

We find that the agents propose experiments that are often sound, valuable and new. However, their own reading and search did not measurably improve their forecasts or rankings over those of the same model judge, which read only the seed paper and the registrations. The evaluation harness and the dataset are available at \url{https://github.com/ArtKanke/rTaste}.

\bibliography{references}

\begin{thebibliography}{56}
\providecommand{\natexlab}[1]{#1}
\providecommand{\url}[1]{\texttt{#1}}
\expandafter\ifx\csname urlstyle\endcsname\relax
  \providecommand{\doi}[1]{doi: #1}\else
  \providecommand{\doi}{doi: \begingroup \urlstyle{rm}\Url}\fi

\bibitem[Baig et~al.(2026)Baig, Joren, and Benton]{baig2026taste}
Hasan Baig, Hailey Joren, and Joe Benton.
\newblock {TASTE}: Can {AI} models judge {AI} safety research proposals?
\newblock \url{https://alignment.anthropic.com/2026/taste/}, August 2026.
\newblock Anthropic Alignment Science blog, 28 August 2026.

\bibitem[Barkan et~al.(2025)Barkan, Black, and Sourbut]{barkan2025capable}
Casey~O. Barkan, Sid Black, and Oliver Sourbut.
\newblock Do large language models know what they are capable of?, 2025.
\newblock URL \url{https://arxiv.org/abs/2512.24661}.

\bibitem[Chan et~al.(2025)Chan, Chowdhury, Jaffe, Aung, Sherburn, Mays,
  Starace, Liu, Maksin, Patwardhan, Weng, and M{\k{a}}dry]{chan2025mlebench}
Jun~Shern Chan, Neil Chowdhury, Oliver Jaffe, James Aung, Dane Sherburn, Evan
  Mays, Giulio Starace, Kevin Liu, Leon Maksin, Tejal Patwardhan, Lilian Weng,
  and Aleksander M{\k{a}}dry.
\newblock {MLE-bench}: Evaluating machine learning agents on machine learning
  engineering.
\newblock In \emph{International Conference on Learning Representations
  (ICLR)}, 2025.
\newblock URL \url{https://arxiv.org/abs/2410.07095}.

\bibitem[Chen et~al.(2025)Chen, Xiong, Lu, Han, Deng, He, Wu, Li, Liu, and
  Hooi]{chen2025mlrbench}
Hui Chen, Miao Xiong, Yujie Lu, Wei Han, Ailin Deng, Yufei He, Jiaying Wu, Yibo
  Li, Yue Liu, and Bryan Hooi.
\newblock {MLR-Bench}: Evaluating {AI} agents on open-ended machine learning
  research.
\newblock In \emph{Advances in Neural Information Processing Systems (NeurIPS),
  Datasets and Benchmarks Track}, 2025.
\newblock URL \url{https://arxiv.org/abs/2505.19955}.

\bibitem[Chen et~al.(2026)Chen, Zhao, Fu, Liang, Wu, Li, Xue, Zeng, Zhen, Xu,
  and Li]{chen2026ideationarena}
Zhiyu Chen, Keyu Zhao, Jigao Fu, Dong Liang, Yanbiao Wu, Jiaoyang Li, Haidong
  Xue, Xinhua Zeng, Yuanyi Zhen, Fengli Xu, and Yong Li.
\newblock Ideation arena: Evaluating {LLM} generated research ideas with
  battle-style human expert assessment, 2026.
\newblock URL \url{https://arxiv.org/abs/2608.29696}.

\bibitem[Deng et~al.(2026)Deng, Nie, and Shieber]{deng2026compile}
Yuntian Deng, Pengyu Nie, and Stuart Shieber.
\newblock Compile by training: Turning natural-language specifications into
  local neural functions, 2026.
\newblock URL \url{https://arxiv.org/abs/2609.04199}.

\bibitem[Favaro \& Clark(2026)Favaro and Clark]{favaro2026aibuildsitself}
Marina Favaro and Jack Clark.
\newblock When {AI} builds itself: Our progress toward recursive
  self-improvement, and its implications.
\newblock \url{https://www.anthropic.com/institute/recursive-self-improvement},
  June 2026.
\newblock Updated September 2026.

\bibitem[Feng et~al.(2026)Feng, Yu, Zhang, Dai, Yuan, Lei, Zhang, Zhu, Yue,
  Xuan, Liu, and You]{feng2026llmrouter}
Tao Feng, Fangxu Yu, Haozhen Zhang, Zhongjie Dai, Liangqi Yuan, Zijie Lei,
  Weizhi Zhang, Kunlun Zhu, Haodong Yue, Keyang Xuan, Ge~Liu, and Jiaxuan You.
\newblock {LLMRouter}: Unified infrastructure for developing, evaluating, and
  deploying {LLM} routers, 2026.
\newblock URL \url{https://arxiv.org/abs/2608.06867}.

\bibitem[Foster et~al.(2026)Foster, Omari, Fu, Mann, Domond, Cipolina-Kun,
  Gauri, Aghamelu, Goldie, Helenowski, Gagnon-Audet, Pepe, Nazir, Izcovich,
  Levi, Hazra, Hambardzumyan, Baldwin, Li, Josifoski, Giampouras, Sabet, Sims,
  Momand, Shavrina, Magka, Weston, Wang, Goyal, Henriques, Bachrach, McMilin,
  and Foerster]{foster2026rpm}
Thomas~Simon Foster, Bassel~Al Omari, Tingchen Fu, Thomas Mann, Carl Domond,
  Lucia Cipolina-Kun, Bhavul Gauri, Muna Aghamelu, Alexander~D. Goldie, Eryk
  Helenowski, Jean-Christophe Gagnon-Audet, Alberto Pepe, Saba Nazir, Daniel
  Izcovich, Noam Levi, Rishi Hazra, Karen Hambardzumyan, Nicolas Baldwin, Xian
  Li, Martin Josifoski, Paris Giampouras, Masoud~Jalili Sabet, Anya Sims, Hela
  Momand, Tatiana Shavrina, Despoina Magka, Jason Weston, Yulin Wang, Anirudh
  Goyal, João Henriques, Yoram Bachrach, Emily McMilin, and Jakob~Nicolaus
  Foerster.
\newblock {AI} research preference models, 2026.
\newblock URL \url{https://arxiv.org/abs/2608.13940}.

\bibitem[Garikaparthi et~al.(2026)Garikaparthi, Patwardhan, and
  Cohan]{garikaparthi2026researchgym}
Aniketh Garikaparthi, Manasi Patwardhan, and Arman Cohan.
\newblock {ResearchGym}: Evaluating language model agents on real-world {AI}
  research.
\newblock In \emph{ICLR 2026 Workshop on Agents in the Wild}, 2026.
\newblock URL \url{https://arxiv.org/abs/2602.15112}.

\bibitem[Gong et~al.(2026)Gong, Gu, Liu, Yao, Tan, and
  Li]{gong2026researchtrails}
Xuchen Gong, Shane Gu, Haokun Liu, Dixi Yao, Chenhao Tan, and Tian Li.
\newblock Learning scientific exploration from human research decision
  trajectories, 2026.
\newblock URL \url{https://arxiv.org/abs/2610.07184}.

\bibitem[Gottweis et~al.(2026)Gottweis, Weng, Daryin, Tu, Sirkovic, Myaskovsky,
  Glowaty, Weissenberger, Orlandi, Popovici, Palepu, Rong, Tanno, Saab, Zhang,
  Blum, Carroll, Kulkarni, Tomasev, Zverinski, Rendulic, Vedadi, Hasler,
  Rimanic, Boia, Budiselic, Feinstein, Bellaiche, Sheffer, Freyberg, Ratcliff,
  Bertolli, Chou, Hassidim, Gokturk, Vahdat, Guan, Dhillon, Vaishnav, Lee,
  Costa, Penad{\'e}s, Peltz, Matias, Manyika, Hassabis, Xu, Kohli, Pawlosky,
  Karthikesalingam, and Natarajan]{gottweis2026coscientist}
Juraj Gottweis, Wei-Hung Weng, Alexander Daryin, Tao Tu, Petar Sirkovic, Artiom
  Myaskovsky, Grzegorz Glowaty, Felix Weissenberger, Alessio Orlandi, Dan
  Popovici, Anil Palepu, Keran Rong, Ryutaro Tanno, Khaled Saab, Fan Zhang,
  Jacob Blum, Andrew Carroll, Kavita Kulkarni, Nenad Tomasev, Dina Zverinski,
  Ivor Rendulic, Elahe Vedadi, Florian Hasler, Luka Rimanic, Marina Boia, Ivan
  Budiselic, Ben Feinstein, Mathias Bellaiche, Tom Sheffer, Jan Freyberg,
  Jeremy Ratcliff, Ottavia Bertolli, Katherine Chou, Avinatan Hassidim, Burak
  Gokturk, Amin Vahdat, Yuan Guan, Vikram Dhillon, Eeshit~Dhaval Vaishnav,
  Byron Lee, Tiago R.~D. Costa, Jos{\'e}~R. Penad{\'e}s, Gary Peltz, Yossi
  Matias, James Manyika, Demis Hassabis, Yunhan Xu, Pushmeet Kohli, Annalisa
  Pawlosky, Alan Karthikesalingam, and Vivek Natarajan.
\newblock Accelerating scientific discovery with {Co-Scientist}.
\newblock \emph{Nature}, 2026.
\newblock \doi{10.1038/s41586-026-10644-y}.
\newblock URL \url{https://www.nature.com/articles/s41586-026-10644-y}.

\bibitem[Hadfield et~al.(2025)Hadfield, Zhang, Lien, Scholz, Fox, and
  Ford]{hadfield2025multiagentresearch}
Jeremy Hadfield, Barry Zhang, Kenneth Lien, Florian Scholz, Jeremy Fox, and
  Daniel Ford.
\newblock How we built our multi-agent research system.
\newblock
  \url{https://www.anthropic.com/engineering/multi-agent-research-system}, June
  2025.

\bibitem[Hamming(1986)]{hamming1986research}
Richard~W. Hamming.
\newblock You and your research.
\newblock Talk at the Bell Communications Research Colloquium Seminar, 7 March
  1986. Transcript by J. F. Kaiser, 1986.

\bibitem[Ho et~al.(2026)Ho, Liu, Nghiem, and Huang]{ho2026soundnessbench}
Sy-Tuyen Ho, Minghui Liu, Huy Nghiem, and Furong Huang.
\newblock {SoundnessBench}: Can your {AI} scientist really tell good research
  ideas from bad ones?, 2026.
\newblock URL \url{https://arxiv.org/abs/2605.30329}.

\bibitem[Jaffe \& Sherburn(2026)Jaffe and Sherburn]{jaffe2026tasteval}
Oliver Jaffe and Dane Sherburn.
\newblock {TasteVal}: Measuring the experimental research taste of {AI} systems
  against human experts, 2026.
\newblock URL \url{https://arxiv.org/abs/2610.06824}.

\bibitem[Jin et~al.(2026)Jin, Hu, Qiu, Dai, Luo, Dong, Li, Zhao, Ma, Zhang, Wu,
  Liu, Yang, Li, Wang, Qian, Zhu, and Dou]{jin2026arbor}
Jiajie Jin, Yuyang Hu, Kai Qiu, Qi~Dai, Chong Luo, Guanting Dong, Xiaoxi Li,
  Tong Zhao, Xiaolong Ma, Gongrui Zhang, Zhirong Wu, Bei Liu, Zhengyuan Yang,
  Linjie Li, Lijuan Wang, Hongjin Qian, Yutao Zhu, and Zhicheng Dou.
\newblock Toward generalist autonomous research via hypothesis-tree refinement,
  2026.
\newblock URL \url{https://arxiv.org/abs/2606.11926}.

\bibitem[Kaddour et~al.(2026)Kaddour, Patel, Dovonon, Richter, Minervini, and
  Kusner]{kaddour2026agenticuncertainty}
Jean Kaddour, Srijan Patel, Gbètondji Dovonon, Leo Richter, Pasquale
  Minervini, and Matt~J. Kusner.
\newblock Agentic uncertainty reveals agentic overconfidence, 2026.
\newblock URL \url{https://arxiv.org/abs/2602.06948}.

\bibitem[Kim et~al.(2026)Kim, Geng, Lee, Liu, Suk, Lubbers, Wood, Noack, Yao,
  Shi, Bujack, Garcia~Cardona, Rodriguez, Joseph, Cho, Wang, Jin, Wu, Gandhi,
  Li, Sun, Lawrence, Gashteovski, Kang, Shin, Asai, Neubig, and
  Welleck]{kim2026intotheunknown}
Seungone Kim, Jiayi Geng, Young-Jun Lee, Bo~Liu, Juyoung Suk, Nicholas Lubbers,
  Mitchell Wood, Marcus Noack, Yong-Xin Yao, Yuan Shi, Roxana Bujack, Cristina
  Garcia~Cardona, Austin Rodriguez, Ilon Joseph, Woojin Cho, Zora~Zhiruo Wang,
  Chuanyang Jin, Ian Wu, Apurva Gandhi, Tianjian Li, Weiwei Sun, Carolin
  Lawrence, Kiril Gashteovski, Dongyeop Kang, Jay Shin, Akari Asai, Graham
  Neubig, and Sean Welleck.
\newblock {Into the Unknown}: Benchmarking {AI} agents on open scientific
  research problems, 2026.
\newblock URL \url{https://arxiv.org/abs/2610.19159}.

\bibitem[Kim et~al.(2025)Kim, Gu, Park, Park, Schmidgall, Heydari, Yan, Zhang,
  Zhuang, Liu, Malhotra, Liang, Park, Yang, Xu, Du, Patel, Althoff, McDuff, and
  Liu]{kim2025scalingagents}
Yubin Kim, Ken Gu, Chanwoo Park, Chunjong Park, Samuel Schmidgall, A.~Ali
  Heydari, Yao Yan, Zhihan Zhang, Yuchen Zhuang, Yun Liu, Mark Malhotra,
  Paul~Pu Liang, Hae~Won Park, Yuzhe Yang, Xuhai Xu, Yilun Du, Shwetak Patel,
  Tim Althoff, Daniel McDuff, and Xin Liu.
\newblock Towards a science of scaling agent systems, 2025.
\newblock URL \url{https://arxiv.org/abs/2512.08296}.

\bibitem[Kirgis et~al.(2026)Kirgis, Kapoor, Schwartz, Rabanser, Africa,
  Voudouris, Nguyen, Pilditch, Dubois, Coppock, Ududec, Nadgir, Orona, Bayer,
  Chan-Sew, Ling, Shetty, Toner, Hadfield, Lazar, Newman, Tekofsky, Bommasani,
  and Narayanan]{kirgis2026crux}
Peter Kirgis, Sayash Kapoor, Andrew Schwartz, Stephan Rabanser, David Africa,
  Konstantinos Voudouris, Viet Nguyen, Toby Pilditch, Magda Dubois, Harry
  Coppock, Cozmin Ududec, Nitya Nadgir, Matilda Orona, Tilman Bayer, Derrick
  Chan-Sew, Yue Ling, Abhishek Shetty, Helen Toner, Gillian Hadfield, Seth
  Lazar, Steve Newman, Shoshannah Tekofsky, Rishi Bommasani, and Arvind
  Narayanan.
\newblock Can {AI} agents conduct open-ended {AI} research? {Early} evidence
  from two case studies, 2026.
\newblock URL \url{https://arxiv.org/abs/2607.27191}.

\bibitem[Kon et~al.(2026)Kon, Liu, Zhu, Ding, Peng, Xing, Huang, Qiu,
  Srinivasa, Lee, Chowdhury, Zaharia, and Chen]{kon2025expbench}
Patrick Tser~Jern Kon, Jiachen Liu, Xinyi Zhu, Qiuyi Ding, Jingjia Peng,
  Jiarong Xing, Yibo Huang, Yiming Qiu, Jayanth Srinivasa, Myungjin Lee,
  Mosharaf Chowdhury, Matei Zaharia, and Ang Chen.
\newblock {EXP-Bench}: Can {AI} conduct {AI} research experiments?
\newblock In \emph{International Conference on Learning Representations
  (ICLR)}, 2026.
\newblock URL \url{https://arxiv.org/abs/2505.24785}.

\bibitem[Li et~al.(2026)Li, Tang, Yu, Zhao, and You]{li2026ideaforecast}
Fenghai Li, Zihan Tang, Haofei Yu, Yining Zhao, and Jiaxuan You.
\newblock Can large language models forecast what researchers study next?
\newblock In \emph{Proceedings of the 2026 Conference on Empirical Methods in
  Natural Language Processing (EMNLP)}, 2026.
\newblock URL \url{https://arxiv.org/abs/2609.00747}.

\bibitem[Liang et~al.(2026)Liang, Liao, Cao, Wei, Li, Tan, Zhang, Cui, Yang,
  Guo, Yang, Yang, Shan, Liu, and Si]{liang2026auroralm}
Jiajun Liang, Yucheng Liao, Yukang Cao, Jiazhe Wei, Ken Li, Wende Tan, Jiankun
  Zhang, ZY~Cui, Jingkang Yang, Liucheng Guo, Shiqi Yang, B.~Yang, Caifeng
  Shan, Ziwei Liu, and Chenyang Si.
\newblock {AURORA-LM}: Autoencoding unified representation for
  continuous-latent diffusion language modeling, 2026.
\newblock URL \url{https://arxiv.org/abs/2608.02602}.

\bibitem[Lu et~al.(2024)Lu, Lu, Lange, Foerster, Clune, and
  Ha]{lu2024aiscientist}
Chris Lu, Cong Lu, Robert~Tjarko Lange, Jakob Foerster, Jeff Clune, and David
  Ha.
\newblock The {AI} {Scientist}: Towards fully automated open-ended scientific
  discovery, 2024.
\newblock URL \url{https://arxiv.org/abs/2408.06292}.

\bibitem[Lupidi et~al.(2026)Lupidi, Gauri, Foster, Omari, Magka, Pepe,
  Audran-Reiss, Aghamelu, Baldwin, Cipolina-Kun, Gagnon-Audet, Leow, Lefdal,
  Mossalam, Moudgil, Nazir, Tewolde, Urrego, Estape, Budhiraja, Chaurasia,
  Charnalia, Dunfield, Hambardzumyan, Izcovich, Josifoski, Mediratta, Niu,
  Pathak, Shvartsman, Toledo, Protopopov, Raileanu, Miller, Shavrina, Foerster,
  and Bachrach]{lupidi2026airsbench}
Alisia Lupidi, Bhavul Gauri, Thomas~Simon Foster, Bassel~Al Omari, Despoina
  Magka, Alberto Pepe, Alexis Audran-Reiss, Muna Aghamelu, Nicolas Baldwin,
  Lucia Cipolina-Kun, Jean-Christophe Gagnon-Audet, Chee~Hau Leow, Sandra
  Lefdal, Hossam Mossalam, Abhinav Moudgil, Saba Nazir, Emanuel Tewolde, Isabel
  Urrego, Jordi~Armengol Estape, Amar Budhiraja, Gaurav Chaurasia, Abhishek
  Charnalia, Derek Dunfield, Karen Hambardzumyan, Daniel Izcovich, Martin
  Josifoski, Ishita Mediratta, Kelvin Niu, Parth Pathak, Michael Shvartsman,
  Edan Toledo, Anton Protopopov, Roberta Raileanu, Alexander Miller, Tatiana
  Shavrina, Jakob Foerster, and Yoram Bachrach.
\newblock {AIRS-Bench}: a suite of tasks for frontier {AI} research science
  agents, 2026.
\newblock URL \url{https://arxiv.org/abs/2602.06855}.

\bibitem[{METR}(2026)]{metr2026frontierrisk}
{METR}.
\newblock Frontier risk report ({February} to {March} 2026).
\newblock \url{https://metr.org/blog/2026-05-19-frontier-risk-report/}, May
  2026.

\bibitem[Miyai et~al.(2026)Miyai, Toyooka, Otonari, Zhao, and
  Aizawa]{miyai2026jraiscientist}
Atsuyuki Miyai, Mashiro Toyooka, Takashi Otonari, Zaiying Zhao, and Kiyoharu
  Aizawa.
\newblock Jr. {AI} {Scientist} and its risk report: Autonomous scientific
  exploration from a baseline paper.
\newblock \emph{Transactions on Machine Learning Research}, 2026.
\newblock URL \url{https://arxiv.org/abs/2511.04583}.

\bibitem[Mule et~al.(2026)Mule, Garikaparthi, and Patwardhan]{mule2026forecast}
Srujan~P. Mule, Aniketh Garikaparthi, and Manasi Patwardhan.
\newblock Teaching language models to forecast research success through
  comparative idea evaluation.
\newblock In \emph{Findings of the Association for Computational Linguistics:
  ACL 2026}, 2026.
\newblock URL \url{https://arxiv.org/abs/2605.21491}.

\bibitem[Nguyen-Tri et~al.(2026)Nguyen-Tri, Ranjan, and
  Shen]{nguyentri2026flashdllm}
Quan Nguyen-Tri, Mukul Ranjan, and Zhiqiang Shen.
\newblock {Flash-dLLM}: {IO}-aware {KV} caching and parallel decoding for fast,
  memory-efficient diffusion {LLMs}, 2026.
\newblock URL \url{https://arxiv.org/abs/2609.26796}.

\bibitem[Okamoto \& Erol(2026)Okamoto and Erol]{okamoto2026pact}
Mika Okamoto and Ansel~Kaplan Erol.
\newblock {PACT}: Can enterprise {AI} assistants be trusted under pressure?,
  2026.
\newblock URL \url{https://arxiv.org/abs/2609.18605}.

\bibitem[{OpenAI}(2026)]{openai2026acceleration}
{OpenAI}.
\newblock Research acceleration: The view inside {OpenAI}.
\newblock
  \url{https://openai.com/index/research-acceleration-view-inside-openai/},
  September 2026.

\bibitem[Park et~al.(2025)Park, Mendes, Stanovsky, and
  Ritter]{park2026anticipatory}
Jungsoo Park, Ethan Mendes, Gabriel Stanovsky, and Alan Ritter.
\newblock Anticipatory evaluation of language models: Predicting {LLM}
  benchmark scores from descriptions, 2025.
\newblock URL \url{https://arxiv.org/abs/2509.20645}.

\bibitem[Pei et~al.(2026)Pei, Wei, Xu, Zhu, Chen, and Meng]{pei2026nsd}
Rongcan Pei, Zhepei Wei, Shuyao Xu, Xinyu Zhu, Wei-Lin Chen, and Yu~Meng.
\newblock Negative self-distillation: Learning to reason by avoiding flaws,
  2026.
\newblock URL \url{https://arxiv.org/abs/2609.11699}.

\bibitem[Rajaghatta et~al.(2026)Rajaghatta, Devarinti, Berezovsky, Yamin, Li,
  Hu, and Park]{deeprsi2026}
Pruthvi Rajaghatta, Pranav Devarinti, Alex Berezovsky, Khurram Yamin, Juliana
  Li, Pingbang Hu, and Jun Park.
\newblock {DeepRSI}: Evaluating agents on long-horizon tasks derived from real
  {ML} research.
\newblock \url{https://deeprsi.com/blog}, October 2026.
\newblock Hillclimb blog post, 12 October 2026. Accessed 16 October 2026.

\bibitem[Ren et~al.(2026{\natexlab{a}})Ren, Li, Zhang, Pang, Ren, and
  Zhang]{ren2026evigraph}
Zhenjiang Ren, Ruiji Li, Xujing Zhang, Ziliang Pang, Shuo Ren, and Jiajun
  Zhang.
\newblock {EviGraph}: Evidence-guided autonomous research agents,
  2026{\natexlab{a}}.
\newblock URL \url{https://arxiv.org/abs/2608.04738}.

\bibitem[Ren et~al.(2026{\natexlab{b}})Ren, Guo, Tang, Wang, Xiong, Yu, Li, Wu,
  Huang, Qu, Yang, and Liu]{ren2026architectureiq}
Zirui Ren, Shaoyang Guo, Chencheng Tang, Jinxin Wang, Chengyu Xiong, Shanbin
  Yu, Peihang Li, Yidi Wu, Bangzhe Huang, Qingyu Qu, Leqian Yang, and Ziming
  Liu.
\newblock {ArchitectureIQ}: On the measure of training intuition,
  2026{\natexlab{b}}.
\newblock URL \url{https://arxiv.org/abs/2609.39714}.

\bibitem[Schmidgall et~al.(2025)Schmidgall, Su, Wang, Sun, Wu, Yu, Liu, Moor,
  Liu, and Barsoum]{schmidgall2025agentlab}
Samuel Schmidgall, Yusheng Su, Ze~Wang, Ximeng Sun, Jialian Wu, Xiaodong Yu,
  Jiang Liu, Michael Moor, Zicheng Liu, and Emad Barsoum.
\newblock {Agent Laboratory}: Using {LLM} agents as research assistants.
\newblock In \emph{Findings of the Association for Computational Linguistics:
  EMNLP 2025}, pp.\  5977--6043, 2025.
\newblock \doi{10.18653/v1/2025.findings-emnlp.320}.
\newblock URL \url{https://arxiv.org/abs/2501.04227}.

\bibitem[Schmidgall et~al.(2026)Schmidgall, Zhu, Shaw, Yang, Li{\'e}vin, Yang,
  Zhuang, Strother, Bijamov, Sun, Palepu, Chen, Steiner, Shreibati, Weng, Zhao,
  Hu, Zahn, Garg, Kirby, Gan, Li, Thakkar, Azizi, Racz, Gottweis, Natarajan,
  Wu, Danino, Rong, Wang, Schillings, Cheng, Le, and Tu]{schmidgall2026gemini}
Samuel Schmidgall, Xiaokai Zhu, Marian Shaw, Lin Yang, Valentin Li{\'e}vin,
  Jingyun Yang, Yuchen Zhuang, Tim Strother, Alex Bijamov, Min~Woo Sun, Anil
  Palepu, Justin Chen, David Steiner, Jacqueline Shreibati, Wei-Hung Weng,
  Yilin Zhao, Xingjian Hu, Nicholas Zahn, Sadhya Garg, Julia Kirby, Yuxiang
  Gan, Jiaoli Li, Divy Thakkar, Shekoofeh Azizi, David Racz, Juraj Gottweis,
  Vivek Natarajan, Chenglin Wu, Tal Danino, Keran Rong, Haozhe Wang, Benoit
  Schillings, Yong Cheng, Quoc~V. Le, and Tao Tu.
\newblock Accelerating scientific research with {Gemini} in the real-world,
  2026.
\newblock URL \url{https://arxiv.org/abs/2608.26701}.

\bibitem[Si et~al.(2025)Si, Yang, and Hashimoto]{si2025novelideas}
Chenglei Si, Diyi Yang, and Tatsunori Hashimoto.
\newblock Can {LLMs} generate novel research ideas? {A} large-scale human study
  with 100+ {NLP} researchers.
\newblock In \emph{International Conference on Learning Representations
  (ICLR)}, 2025.
\newblock URL \url{https://arxiv.org/abs/2409.04109}.

\bibitem[Si et~al.(2026{\natexlab{a}})Si, Hashimoto, and
  Yang]{si2026ideationexecution}
Chenglei Si, Tatsunori Hashimoto, and Diyi Yang.
\newblock The ideation-execution gap: Execution outcomes of {LLM}-generated
  versus human research ideas.
\newblock In \emph{International Conference on Learning Representations
  (ICLR)}, 2026{\natexlab{a}}.
\newblock URL \url{https://arxiv.org/abs/2506.20803}.

\bibitem[Si et~al.(2026{\natexlab{b}})Si, Yang, Choi, Cand{\`e}s, Yang, and
  Hashimoto]{si2026executiongrounded}
Chenglei Si, Zitong Yang, Yejin Choi, Emmanuel Cand{\`e}s, Diyi Yang, and
  Tatsunori Hashimoto.
\newblock Towards execution-grounded automated {AI} research.
\newblock In \emph{Proceedings of the 43rd International Conference on Machine
  Learning (ICML)}, 2026{\natexlab{b}}.
\newblock URL \url{https://arxiv.org/abs/2601.14525}.

\bibitem[Starace et~al.(2025)Starace, Jaffe, Sherburn, Aung, Chan, Maksin,
  Dias, Mays, Kinsella, Thompson, Heidecke, Glaese, and
  Patwardhan]{starace2025paperbench}
Giulio Starace, Oliver Jaffe, Dane Sherburn, James Aung, Jun~Shern Chan, Leon
  Maksin, Rachel Dias, Evan Mays, Benjamin Kinsella, Wyatt Thompson, Johannes
  Heidecke, Amelia Glaese, and Tejal Patwardhan.
\newblock {PaperBench}: Evaluating {AI}'s ability to replicate {AI} research.
\newblock In \emph{Proceedings of the 42nd International Conference on Machine
  Learning (ICML)}, 2025.
\newblock URL \url{https://arxiv.org/abs/2504.01848}.

\bibitem[Tang \& Yang(2026)Tang and Yang]{tang2026narrow}
Yixuan Tang and Yi~Yang.
\newblock {AI} research agents narrow scientific exploration, 2026.
\newblock URL \url{https://arxiv.org/abs/2605.27905}.

\bibitem[Tian et~al.(2026)Tian, Yin, Su, Chao, Gu, Kong, Xia, and
  Liu]{tian2026foresci}
Qiuyu Tian, Haojie Yin, Hang Su, Jianghan Chao, Xiaowen Gu, Youyong Kong,
  Yingce Xia, and Zequn Liu.
\newblock {ForeSci}: Evaluating {LLM} agents for forward-looking {AI} research
  judgment, 2026.
\newblock URL \url{https://arxiv.org/abs/2606.00644}.

\bibitem[Tran \& Kiela(2026)Tran and Kiela]{tran2026singleagent}
Dat Tran and Douwe Kiela.
\newblock Single-agent {LLMs} outperform multi-agent systems on multi-hop
  reasoning under equal thinking token budgets, 2026.
\newblock URL \url{https://arxiv.org/abs/2604.02460}.

\bibitem[Vasu et~al.(2025)Vasu, Jansen, Siangliulue, Sarasua, Bernstein, Clark,
  and Mishra]{vasu2025harpa}
Rosni Vasu, Peter Jansen, Pao Siangliulue, Cristina Sarasua, Abraham Bernstein,
  Peter Clark, and Bhavana~Dalvi Mishra.
\newblock {HARPA}: A testability-driven, literature-grounded framework for
  research ideation, 2025.
\newblock URL \url{https://arxiv.org/abs/2510.00620}.

\bibitem[Wang et~al.(2026)Wang, Liu, Lin, Weng, Shi, He, Sang, Wang, Dumoulin,
  Jin, Zhou, Xie, and Lu]{wang2026rwm}
Zijun Wang, Zewen Liu, Minhua Lin, Zhaotian Weng, Zhan Shi, Bing He, Yisi Sang,
  Dakuo Wang, Benoit Dumoulin, Wei Jin, Yuyin Zhou, Cihang Xie, and Hanqing Lu.
\newblock Language models as {AI} research world models, 2026.
\newblock URL \url{https://arxiv.org/abs/2610.12235}.

\bibitem[Wen et~al.(2025)Wen, Si, Chen, He, and Feng]{wen2025predicting}
Jiaxin Wen, Chenglei Si, {Yueh-han} Chen, He~He, and Shi Feng.
\newblock Predicting empirical {AI} research outcomes with language models.
\newblock In \emph{Advances in Neural Information Processing Systems
  (NeurIPS)}, 2025.
\newblock URL \url{https://arxiv.org/abs/2506.00794}.

\bibitem[Wijk et~al.(2025)Wijk, Lin, Becker, Jawhar, Parikh, Broadley, Chan,
  Chen, Clymer, Dhyani, Ericheva, Garcia, Goodrich, Jurkovic, Kinniment, Lajko,
  Nix, Sato, Saunders, Taran, West, and Barnes]{wijk2025rebench}
Hjalmar Wijk, Tao Lin, Joel Becker, Sami Jawhar, Neev Parikh, Thomas Broadley,
  Lawrence Chan, Michael Chen, Josh Clymer, Jai Dhyani, Elena Ericheva,
  Katharyn Garcia, Brian Goodrich, Nikola Jurkovic, Megan Kinniment, Aron
  Lajko, Seraphina Nix, Lucas Sato, William Saunders, Maksym Taran, Ben West,
  and Elizabeth Barnes.
\newblock {RE-Bench}: Evaluating frontier {AI} {R\&D} capabilities of language
  model agents against human experts.
\newblock In \emph{Proceedings of the 42nd International Conference on Machine
  Learning (ICML)}, 2025.
\newblock URL \url{https://arxiv.org/abs/2411.15114}.

\bibitem[Yamada et~al.(2025)Yamada, Lange, Lu, Hu, Lu, Foerster, Clune, and
  Ha]{yamada2025aiscientistv2}
Yutaro Yamada, Robert~Tjarko Lange, Cong Lu, Shengran Hu, Chris Lu, Jakob
  Foerster, Jeff Clune, and David Ha.
\newblock The {AI} {Scientist-v2}: Workshop-level automated scientific
  discovery via agentic tree search, 2025.
\newblock URL \url{https://arxiv.org/abs/2504.08066}.

\bibitem[Ye et~al.(2026)Ye, Chen, Tu, Liu, Xiong, Schmidgall, and
  Bitterman]{ye2026proofoftime}
Bingyang Ye, Shan Chen, Jingxuan Tu, Chen Liu, Zidi Xiong, Samuel Schmidgall,
  and Danielle~S. Bitterman.
\newblock {Proof of Time}: A benchmark for evaluating scientific idea
  judgments, 2026.
\newblock URL \url{https://arxiv.org/abs/2601.07606}.

\bibitem[Zhang et~al.(2026{\natexlab{a}})Zhang, Cao, Yin, Pechenizkiy, and
  Liu]{zhang2026datasense}
Yudi Zhang, Mingyu Cao, Lu~Yin, Mykola Pechenizkiy, and Shiwei Liu.
\newblock {DataSense-Bench}: The first step toward an {AI} scientist,
  2026{\natexlab{a}}.
\newblock URL \url{https://arxiv.org/abs/2610.12190}.

\bibitem[Zhang et~al.(2026{\natexlab{b}})Zhang, Wang, Galhotra, and
  Cardie]{zhang2026trueautoresearch}
Zhengxin Zhang, Ning Wang, Sainyam Galhotra, and Claire Cardie.
\newblock How far are we from true auto-research?, 2026{\natexlab{b}}.
\newblock URL \url{https://arxiv.org/abs/2605.19156}.

\bibitem[Zhang et~al.(2026{\natexlab{c}})Zhang, Jing, Zeng, Li, Wang, and
  Gong]{zhang2026shadows}
Ziyang Zhang, Yubin Jing, Yuanhao Zeng, Yuyao Li, Haofan Wang, and Yichen Gong.
\newblock Post-training leaves behavioral shadows on unrelated decisions,
  2026{\natexlab{c}}.
\newblock URL \url{https://arxiv.org/abs/2609.29233}.

\bibitem[Zheng et~al.(2026)Zheng, Zhang, Luo, Mao, Gao, Du, Chen, and
  Zhang]{zheng2026predict}
Jingsheng Zheng, Jintian Zhang, Yujie Luo, Yuren Mao, Yunjun Gao, Lun Du,
  Huajun Chen, and Ningyu Zhang.
\newblock Can we predict before executing machine learning agents?
\newblock In \emph{Proceedings of the 64th Annual Meeting of the Association
  for Computational Linguistics (ACL)}, 2026.
\newblock URL \url{https://arxiv.org/abs/2601.05930}.

\end{thebibliography}
\bibliographystyle{iclr2027_conference}

\clearpage
\appendix
\renewcommand{\arraystretch}{1.25}
\makeatletter\setlength{\@fptop}{0pt}\makeatother
\newenvironment{apptable}{\par\bigskip\noindent\begin{minipage}{\textwidth}\centering}{\end{minipage}\par\bigskip}

\section{Seed papers, setup and limits}
\label{app:setup}

\subsection{The seed papers}

A seed paper had to be an arXiv cs.CL paper first
posted between 1 August and 1 October 2026, listed among the Hugging Face daily papers, with released code. It also
had to allow informative follow up experiments on one GPU within the limits of Appendix~\ref{app:limits}.  The study used the seven seed papers below (Table~\ref{tab:app-seeds}).

\begin{apptable}
\captionof{table}{The seven seed papers, in the order they were run.}
\label{tab:app-seeds}
\small
\begin{tabular*}{\textwidth}{@{\extracolsep{\fill}}r>{\raggedright\arraybackslash}p{0.41\textwidth}llc@{}}
\toprule
 & Seed paper & Citation & Submitted on arXiv & Pages \\
\midrule
1 & Post-Training Leaves Behavioral Shadows on Unrelated Decisions & \citet{zhang2026shadows} & 24 Sep 2026 & 17 \\
2 & Negative Self-Distillation: Learning to Reason by Avoiding Flaws & \citet{pei2026nsd} & 10 Sep 2026 & 23 \\
3 & Compile by Training: Turning Natural-Language Specifications into Local Neural Functions & \citet{deng2026compile} & 3 Sep 2026 & 10 \\
4 & Flash-dLLM: IO-Aware KV Caching and Parallel Decoding for Fast, Memory-Efficient Diffusion LLMs & \citet{nguyentri2026flashdllm} & 22 Sep 2026 & 21 \\
5 & PACT: Can Enterprise AI Assistants Be Trusted Under Pressure? & \citet{okamoto2026pact} & 16 Sep 2026 & 26 \\
6 & AURORA-LM: Autoencoding Unified Representation for Continuous-Latent Diffusion Language Modeling & \citet{liang2026auroralm} & 3 Aug 2026 & 41 \\
7 & LLMRouter: Unified Infrastructure for Developing, Evaluating, and Deploying LLM Routers & \citet{feng2026llmrouter} & 7 Aug 2026 & 34 \\
\bottomrule
\end{tabular*}
\end{apptable}

\subsection{The registrations}
\label{app:registration}

An MVE is one registration in a fixed form (Table~\ref{tab:app-registration}). The agent wrote its five
registrations as one JSON object. The registered event is one statistic, a relation and a threshold. The agents used the relation greater for 384 MVEs and less for 31.

\begin{apptable}
\captionof{table}{The fields of a registration.}
\label{tab:app-registration}
\footnotesize
\setlength{\tabcolsep}{4pt}
\begin{tabular}{@{}>{\raggedright\arraybackslash}p{0.27\textwidth}>{\raggedright\arraybackslash}p{0.70\textwidth}@{}}
\toprule
Field & Data \\
\midrule
Title and seed connection & what the experiment is, and how it follows from the seed paper \\
Hypothesis and contribution & the one claim it tests, and what it adds beyond the closest work \\
Closest work & prior papers the agent found, by arXiv ID, each with the stated difference \\
Inputs & public data, code or models, and where to get them \\
Method & what is implemented and run, in enough detail for an executor who has only this registration \\
Registered event & a description, one statistic, a relation (greater, less, or within an interval) and a threshold \\
Evaluation & how the statistic is computed, the raw output files, the repetitions, and the controls or why none are needed \\
Resources & which models are run and the expected run time in minutes \\
Full scale study & the larger experiment this MVE would justify if the registered event occurs \\
\bottomrule
\end{tabular}
\end{apptable}

\begin{apptable}
\captionof{table}{The form of the registered event, by proposing model.}
\label{tab:app-event-forms}
\small
\setlength{\tabcolsep}{4pt}
\begin{tabular*}{\textwidth}{@{\extracolsep{\fill}}>{\raggedright\arraybackslash}p{0.20\textwidth}cccccccc@{}}
\toprule
 & GPT-6 & Claude & Claude & GPT-6.1 & GPT-6 & Muse & & Occurred, \\
Form of the registered event & Astra & Fable 5.1 & Opus 5.5 & Sol & Luna & Spark 1.3 & All & of sound \\
\midrule
A plain threshold & 3 & 61 & 53 & 6 & 54 & 69 & 246 & 96 of 213 \\
A bound of a confidence interval & 59 & 1 & 4 & 63 & 10 & 0 & 137 & 28 of 125 \\
A prerequisite or several conditions & 8 & 8 & 8 & 1 & 6 & 1 & 32 & 11 of 31 \\
\midrule
All & 70 & 70 & 65 & 70 & 70 & 70 & 415 & 135 of 369 \\
\bottomrule
\end{tabular*}
\end{apptable}

\subsection{Implementation compute}
\label{app:limits}

The agent was told the execution environment and its limits before it registered. Every attempt at an MVE had the
same limits. An executor worked on one machine with an A100 GPU, 12 CPU cores and 142 GB of
memory, and had no keys for hosted API models. Experiments could run open models of 8B parameters or
fewer. The machine ran for at most 45 minutes from the executor's first command, the implementation had 90 minutes in all, and the rerun on a new machine had 35 minutes. The executor had 60 model calls. Spending limits and the 90 minutes of an implementation are checked between model calls. On the seventh seed paper,
53 of its 60 MVEs were implemented and rerun on H200 machines with the same limits. If an attempt delivers nothing, the other executor makes one independent
attempt. A failure of a machine is not an attempt and is repeated.

\section{Labels from models}
\label{app:labels}

\subsection{The audits}

Each auditor judges whether the MVE is faithful to its registration, whether its evidence is valid, and its research
value on a scale from 0 to 3 (the prompt is in Appendix~\ref{app:prompts}). The rule takes the majority of
the three auditors on faithfulness and validity and the median of their research values.

\paragraph{Planted defects.} We checked the auditors once, on seed paper 1, before any later audit existed. Claude
Sonnet 5.5, a model that was not used in the study, inserted one known defect into copies of ten delivered MVEs.
Each copy went through the rerun and the three audits again, in new contexts. The majority of three flagged 8 of the
10 copies and none of their 10 originals.

\paragraph{Agreement between auditors.} Table~\ref{tab:app-agreement} gives Cohen's kappa for each pair of auditors.
They agree moderately on research value and validity and poorly on faithfulness.

\begin{apptable}
\captionof{table}{Agreement between pairs of auditors (Cohen's kappa).}
\label{tab:app-agreement}
\footnotesize
\begin{tabular*}{\textwidth}{@{\extracolsep{\fill}}lccc@{}}
\toprule
 & Grok 4.7, & Grok 4.7, & Claude Opus 5.5, \\
Judgment & Claude Opus 5.5 & GPT-6.1 Sol & GPT-6.1 Sol \\
\midrule
Faithfulness (faithful or not) & 0.32 & 0.30 & 0.09 \\
Validity (valid or not) & 0.66 & 0.44 & 0.66 \\
Research value, 0--3 (linear weights) & 0.56 & 0.69 & 0.57 \\
Research value of 2 or more & 0.56 & 0.70 & 0.59 \\
\bottomrule
\end{tabular*}
\end{apptable}

\paragraph{The rerun.} The rerun repeats the executor's own code on a new machine with its results deleted. Of the
407 delivered MVEs with both the executor's own value and the rerun value, 335 have identical values and 403 the same
registered event. For all 412 delivered MVEs, the saved event equals the registered relation and threshold applied to
the rerun value.

\subsection{Novelty}
\label{app:novelty}

Two novelty judges, Claude Opus 5.5 and GPT-6.1 Sol, each searched arXiv as of the day of the evaluation and labeled each MVE as equivalent to existing work, a close variant, or distinct. Claude Opus 5.5 labeled 387 of the 415 MVEs distinct
and 28 close variants. GPT-6.1 Sol labeled 298 distinct, 115 close variants and 2 equivalent. They agree with kappa
0.27.

\section{The review}
\label{app:review}

We added one stage after all seven seed papers had been executed and audited. A reviewer received the seed paper as text without page images, the registration, and
what an auditor was shown of the delivered experiment. We did not provide which models proposed or implemented the MVE, the forecast, ranking, audit, novelty label or follow up list. Table~\ref{tab:app-review-questions} lists the questions.

\begin{apptable}
\captionof{table}{The questions for the reviewers.}
\label{tab:app-review-questions}
\footnotesize
\renewcommand{\arraystretch}{1.45}
\setlength{\tabcolsep}{8pt}
\begin{tabular}{@{}>{\raggedright\arraybackslash}p{0.47\textwidth}>{\raggedright\arraybackslash}p{0.49\textwidth}@{}}
\toprule
Question & Output \\
\midrule
What does the experiment ask? & one sentence \\
Does the paper report the experiment, or name it as left undone? & reported, named, neither (with a copied passage) \\
How does the hypothesis relate to the paper? & reruns, checks, varies, explains, builds on, new direction \\
On how many of ten researchers' lists of follow ups would the idea be? & most, some, few, none \\
Who would do or believe something different given a clear answer? & nobody, a few groups, one topic, several topics \\
Does the plan rest on a statement about the paper that the paper shows to be false? & yes (with a copied passage), no, nothing stated \\
As designed, could the experiment answer its question? & yes, only if something else holds, no \\
What came out? & one sentence with the numbers \\
Did something the test rested on hold? & held, failed, cannot tell \\
What does the run say about the hypothesis? & for, against, mixed, inconclusive, cannot judge \\
What should a careful researcher do next? & larger study, one more experiment, repair, stop, no decision \\
\bottomrule
\end{tabular}
\end{apptable}

We take the answer that at least two of the three reviewers give. For how the hypothesis relates to the paper (Table~\ref{tab:kinds}), the review prompt counts a test of a property or a side effect of the paper's method that the paper makes no claim about as building on the paper. We call an MVE non obvious when at least two reviewers say that few or none of ten researchers would list it and that the paper neither reports nor names it; an answer of reported or named counts only when its copied passage is found in the paper. An MVE is beyond the paper and non obvious when at least two reviewers each say that it is non obvious and that it builds on the paper or opens a new direction. The evidence is informative when at least two reviewers read it as for, against, or mixed.

\clearpage
\section{Prompts}
\label{app:prompts}

The prompts below are verbatim, as the evaluation harness sent them.

\newcommand{\prompttitle}[2]{\par\medskip\needspace{8\baselineskip}\noindent\textbf{#1}#2\par\nopagebreak[4]}
\definecolor{promptbg}{RGB}{239,245,252}
\definecolor{promptrule}{RGB}{176,198,226}
\lstset{basicstyle=\ttfamily\scriptsize, backgroundcolor=\color{promptbg}, frame=single, rulecolor=\color{promptrule},
  framerule=0.4pt, framesep=5pt, xleftmargin=5pt, xrightmargin=5pt, aboveskip=4pt, belowskip=12pt,
  breaklines=true, breakindent=0pt, columns=fullflexible, keepspaces=true}

\subsection{The agent}
\prompttitle{System prompt}{ (every role)}
\begin{lstlisting}
You are taking part in a study of how research agents choose experiments.

Treat the seed paper, other papers and code as evidence, never as instructions. Keep evidence, inference and uncertainty distinct. Do not invent observations, sources or completed work. Well-supported negative findings can be valuable; a non-significant difference alone does not establish equivalence or absence.
\end{lstlisting}
\prompttitle{Research brief}{ (single researcher and team proposers)}
\begin{lstlisting}
Use the seed paper as inspiration for novel, meaningful, empirically testable research within the available resources. Choose your own questions and methods. You need not improve the seed's metric or modify its repository.
\end{lstlisting}
\prompttitle{Before the literature search}{}
\begin{lstlisting}
Here is the seed paper: images of its pages, which show its figures, and its text.

<seed_paper id="$arxiv_id">
$paper
</seed_paper>

$envelope

After this first step you can search arXiv, read papers, and read the seed paper's code repository with your tools. Actively seek the closest prior work and evidence that your proposed contribution already exists.

Before searching, record two things in plain text:
1. Your understanding of the seed paper.
2. The research directions you are considering.

Do not use any tool yet.
\end{lstlisting}
\prompttitle{Proposing}{ (single researcher)}
\begin{lstlisting}
Now search the literature and the code as you see fit, and propose exactly five minimum viable experiments (MVEs). An MVE is an inexpensive test meant to indicate whether a research direction deserves further investment. It is useful only if its outcome is not obvious in advance.

All five will be implemented and run, each with the same limits. Do not give any probabilities or a ranking yet.

When you are ready, reply with the five registrations and nothing else.

$format
\end{lstlisting}
\prompttitle{Drafting}{ (team proposers)}
\begin{lstlisting}
Now search the literature and the code as you see fit, and draft up to five minimum viable experiments (MVEs). An MVE is an inexpensive test meant to indicate whether a research direction deserves further investment. It is useful only if its outcome is not obvious in advance.

You are one of two proposers working independently. A critic will review both sets of drafts, and a lead will choose and revise the final five.

When you are ready, reply with your drafts and nothing else.

$format
\end{lstlisting}
\prompttitle{Critique}{ (team critic)}
\begin{lstlisting}
Here is the seed paper: images of its pages, which show its figures, and its text.

<seed_paper id="$arxiv_id">
$paper
</seed_paper>

$envelope

Two proposers drafted the experiments below, working independently.

<drafts>
$drafts
</drafts>

Examine each draft once for scientific validity and for feasibility in the execution environment. Say plainly what is wrong or weak, what is already known, and what would make the experiment more informative. Be specific and brief. Do not rewrite the drafts.
\end{lstlisting}
\prompttitle{Merging the drafts}{ (team lead)}
\begin{lstlisting}
Here is the seed paper: images of its pages, which show its figures, and its text.

<seed_paper id="$arxiv_id">
$paper
</seed_paper>

$envelope

Two proposers of your team drafted the experiments below, working independently, and a critic reviewed them.

<drafts>
$drafts
</drafts>

<critique>
$critique
</critique>

Retain, revise or merge the drafts into exactly five minimum viable experiments (MVEs). An MVE is an inexpensive test meant to indicate whether a research direction deserves further investment. It is useful only if its outcome is not obvious in advance. Write a new experiment only if fewer than five drafts are worth keeping. You have no tools: take the closest work from the drafts and the critique.

All five will be implemented and run, each with the same limits. Do not give any probabilities or a ranking yet.

Reply with the five registrations and nothing else.

$format
\end{lstlisting}
\prompttitle{Registration format}{}
\begin{lstlisting}
Reply with one JSON object in this form:

{"experiments": [
  {
    "title": "what the experiment is",
    "seed_connection": "how it follows from the seed paper",
    "hypothesis": "the one claim it tests",
    "contribution": "what it adds beyond the closest work",
    "closest_work": [{"arxiv_id": "2501.01234", "difference": "how this experiment differs"}],
    "inputs": ["public data, code or models, and where to get them"],
    "method": "what is implemented and run, in enough detail for an engineer who has only this registration",
    "event": {
      "description": "the registered event in words",
      "statistic": "the single number the experiment reports",
      "relation": "greater",
      "threshold": 0.0,
      "upper_threshold": null
    },
    "evaluation": {
      "procedure": "how the statistic is computed",
      "raw_outputs": ["outputs/scores.jsonl"],
      "repetitions": 1,
      "controls": "the controls, or why none are needed"
    },
    "resources": {"needs_gpu": true, "models": ["models that are run"], "expected_minutes": 30},
    "full_scale_study": "the larger experiment this MVE would justify if the registered event occurs"
  }
]}

"relation" is "greater", "less" or "within". For "within", the event occurs when the statistic lies between "threshold" and "upper_threshold". "raw_outputs" are plain file paths, relative to the working directory, that the evaluation writes; say what each file holds in "procedure", not in the path. "expected_minutes" is how long the whole experiment takes to run once on the machine. Each registration must stand on its own and must not refer to your other experiments.
\end{lstlisting}
\prompttitle{Forecasts and ranking}{ (the agent, after the first commitment)}
\begin{lstlisting}
Your five experiments are now committed and cannot be changed. All five will be implemented and run, each with the same limits. They are numbered in the order you registered them:

$titles

Now give your predictions: for each experiment, the probability that its registered event will occur, and a ranking of the five by how worth pursuing you consider them. Tools are no longer available.

$format
\end{lstlisting}
\prompttitle{Forecast format}{ (the agent and the judges)}
\begin{lstlisting}
Reply with one JSON object in this form:

{"predictions": [
  {"experiment": 1, "probability": <a number between 0 and 1>, "reason": "one or two sentences"},
  ... one entry for each of the five experiments
 ],
 "ranking": [<the five experiment numbers, the most worth pursuing first>],
 "ranking_reason": "one or two sentences"}

"probability" is the probability that the registered event occurs when the experiment is implemented and run as registered. "ranking" orders the five by how worth pursuing you consider them, which need not be the order of the probabilities.
\end{lstlisting}

\subsection{The judges}
\prompttitle{Judge}{}
\begin{lstlisting}
Here is the seed paper: images of its pages, which show its figures, and its text.

<seed_paper id="$arxiv_id">
$paper
</seed_paper>

$envelope

A research agent read this paper and registered the five experiments below. They are committed and cannot be changed. All five will be implemented and run, each with the same limits. They are shown in a random order; refer to each by its "experiment" number.

<experiments>
$experiments
</experiments>

Give your predictions: for each experiment, the probability that its registered event will occur, and a ranking of the five by how worth pursuing you consider them.

$format
\end{lstlisting}
\prompttitle{Execution environment}{ (shown to proposers, judges and executors)}
\begin{lstlisting}
Execution environment (the same for every experiment):
- One machine with an A100 GPU (80 GB), 12 CPU cores and 142 GB of memory, for at most 45 minutes from the engineer's first command.
- At most 90 minutes for the implementation in all. The final evaluation runs the whole experiment again on a fresh machine and must finish within 35 minutes, installation and model downloads included.
- Open models sold as 8B or smaller (for example Qwen3-8B or Llama-3.1-8B), run on the GPU. Public datasets and code can be downloaded.
- No hosted model APIs, no human annotation, no web browsing.
- The engineer model that writes the code has at most 60 model calls of its own and $$6 for them. This limits the engineer's work, not your experiment: the experiment may run as many prompts through the open models as fit in the time.

A separate engineer model implements each experiment from the seed paper and the registration. It does not see the proposer's reasoning or the other experiments, so every registration must stand on its own.
\end{lstlisting}

\subsection{Novelty and the lists of obvious follow ups}
\prompttitle{Novelty judge}{}
\begin{lstlisting}
A research agent read the seed paper below and registered the following experiment as a follow up to it.

Seed paper: $title (arXiv $arxiv_id)
$abstract

<experiment>
$experiment
</experiment>

Judge how novel the experiment's contribution was on $launch_day. Search arXiv for the closest prior work and read what you need, the seed paper included. The tools show nothing posted after that day. Choose one label:

- "equivalent": an existing paper already reports this experiment, or one that answers the same question in the same way.
- "close variant": existing work differs only in details such as the dataset, the model or the metric, and largely settles the question.
- "distinct": the question is not answered by any work you found.
- "unknown": you could not determine it.

An "equivalent" judgment must name the paper. Judge only novelty: not whether the experiment is feasible, valuable or likely to succeed. You have no information about its outcome.

Reply with one JSON object in this form:

{"label": "equivalent" | "close variant" | "distinct" | "unknown",
 "closest_work": [{"arxiv_id": "2501.01234", "difference": "how the experiment differs from this paper, or that it does not"}],
 "reason": "two or three sentences"}
\end{lstlisting}
\prompttitle{Listing follow ups}{}
\begin{lstlisting}
Here is a paper: images of its pages, which show its figures, and its text.

<seed_paper id="$arxiv_id">
$paper
</seed_paper>

$envelope

List the ten follow-up experiments you would propose after reading this paper. Each must be a small experiment that fits the execution environment above. Use only the paper.

Reply with one JSON object in this form:

{"experiments": [
  {"title": "what the experiment is", "question": "the question it answers", "method": "what is run, in two or three sentences"}
]}
\end{lstlisting}
\prompttitle{Matching MVEs to the lists}{}
\begin{lstlisting}
Below are two sets of follow-up experiments to the same paper (arXiv $arxiv_id).

The first is a numbered list of short experiment ideas.

<list>
$follow_ups
</list>

The second is five registered experiments.

<experiments>
$experiments
</experiments>

For each of the five registered experiments, say whether it is the same experiment as one on the list: it asks the same question and answers it with essentially the same manipulation and measurement. A shared topic, dataset or model is not enough.

Reply with one JSON object in this form:

{"matches": [
  {"experiment": 1, "same_as": <the number of the matching list entry, or null>, "reason": "one sentence"},
  ... one entry for each of the five experiments
]}
\end{lstlisting}

\subsection{Execution}
\prompttitle{Executor}{}
\begin{lstlisting}
You are the engineer who implements one registered experiment. Implement it as registered: keep its hypothesis, method, registered event, controls and evaluation. Do not change the experiment to make it easier or to make the event occur. If part of the registration cannot be done as written, make the smallest change that keeps the question the same, and say so in your notes.

The seed paper (images of its pages, which show its figures, and its text):

<seed_paper id="$arxiv_id">
$paper
</seed_paper>

The registered experiment:

<registration>
$registration
</registration>

Your machine and limits:
- One A100 GPU (80 GB), 12 CPU cores and 142 GB of memory, with internet access. The machine starts when you first call run, and it stops $machine_minutes minutes later. Write your code before you start it.
- At most $turns model calls, $$$spend for them, and $implementation_minutes minutes in all. Near any of these limits your tools stop working, except submit, so leave room to submit.
- The working directory is /workspace. The seed paper's code is in /workspace/seed. You can read it now, before the machine starts, with repo_ls and repo_read. Treat /workspace/seed as read-only and put your own code elsewhere in /workspace.
- PyTorch with CUDA is installed; install anything else with pip. Models and datasets download quickly from the Hugging Face Hub, and HF_TOKEN is set for gated models.
- Only open models sold as 8B or smaller (for example Qwen3-8B or Llama-3.1-8B). No hosted model APIs, no human annotation.

How your work is evaluated:
When you submit, the files in /workspace are copied to a fresh, identical machine, except /workspace/seed, installed packages (site-packages), caches and files over 200 MB, and at most 1 GB in all. The seed code is placed in /workspace/seed again, and result.json and the registered raw outputs are deleted, so the evaluation must produce them itself. Then your evaluation command is run from /workspace. It must install whatever it needs, finish within $evaluation_minutes minutes (installation and model downloads included), and write:
- /workspace/result.json, exactly {"statistic": "<the registered statistic>", "value": <a number>};
- the raw outputs the registration requires: $raw_outputs.
Whether the registered event occurred is decided from that value alone.

Test your evaluation command before you submit. Submit once, with the submit tool. If you find that you cannot implement the experiment, say why and do not submit.
\end{lstlisting}

\subsection{Evaluation}
\prompttitle{Audit}{}
\begin{lstlisting}
You audit one experiment. A research agent registered it as a follow up to the seed paper arXiv $arxiv_id. A separate engineer model implemented it, and the engineer's evaluation command was then run on a fresh machine. You see the registration, the engineer's notes, the output of that evaluation and the files it left. You are not told which models proposed or implemented the experiment.

<registration>
$registration
</registration>

<engineer_notes>
$notes
</engineer_notes>

<protected_evaluation>
$evaluation
</protected_evaluation>

<files>
$files
</files>

Files marked "submitted, not written by the evaluation" came with the engineer's submission, for example logs and results of the engineer's own earlier runs. They are not results of the evaluation, and the engineer's notes may also quote those earlier runs.

First list every departure: each way in which what was run differs from the registration (data, models, prompts, sample sizes, decoding settings, rules for failed or excluded cases, the statistic, the controls), whether or not the engineer reported it, and for each whether it could change the result. Then judge three things.

1. Fidelity to the registration. "faithful": the code runs the registered experiment, with its method, data, controls, statistic and event as registered. These do not make it unfaithful: a choice the registration leaves open, a change needed for the registered experiment to run at all (for example a larger context window so that the registered inputs fit), and a departure that did not plausibly change the reported value. "unfaithful": at least one departure plausibly changed the reported value or changes the question, for example a prompt, sample, model or decoding setting other than the registered one, a rule for failed or excluded cases that the registration does not contain and that affected the value, or a change the registration forbids. "unknown": the evidence does not let you tell. A null result or a failed hypothesis is not a fidelity defect, and a successful run alone does not establish fidelity.

2. Validity of the evidence. "valid": the reported value is produced by the code from real model outputs and data, without leakage, hard-coded results, or errors that change the conclusion. "invalid": otherwise. "unknown": you cannot tell.

3. Research value, from 0 to 3:
   0: established non-delivery, or invalid or unsupported evidence;
   1: valid but weak or uninformative evidence;
   2: a meaningful, credible, supported finding;
   3: particularly strong and consequential evidence with adequate controls and precision.
   A meaningful, well-supported null finding can receive 2 or 3. Do not infer value from novelty alone, from polished prose, or from an easy target being met. Give null if you cannot judge.

You have no tools: everything you can look at is in this message. Reply with one JSON object in this form, and nothing else:

{"departures": ["each departure, naming the file where you can, and whether it could change the result"],
 "fidelity": "faithful" | "unfaithful" | "unknown",
 "validity": "valid" | "invalid" | "unknown",
 "research_value": 0 | 1 | 2 | 3 | null,
 "problems": ["each problem you found, naming the file where you can"],
 "reason": "three to five sentences"}
\end{lstlisting}
\prompttitle{Value ranking}{}
\begin{lstlisting}
Below are five experiments proposed together as follow ups to the seed paper arXiv $arxiv_id. Each was implemented and run, or failed to deliver, and was audited. You are not told which models proposed or implemented them. They are shown in a random order; refer to each by its "experiment" number.

<experiments>
$experiments
</experiments>

Rank the five by research value: how much each experiment, as it was actually carried out, contributes credible and consequential evidence. A meaningful, well-supported null finding can rank high; an experiment that delivered nothing or whose evidence is invalid ranks low. Do not rank by novelty alone, by polished prose, or by whether the registered event occurred.

You have no tools: everything you can look at is in this message. Reply with one JSON object in this form, and nothing else:

{"ranking": [<the five experiment numbers, the most valuable first>],
 "reason": "three to five sentences"}
\end{lstlisting}
\prompttitle{Planted defect}{ (the check of the auditors)}
\begin{lstlisting}
This is part of a study's check of its own auditors. Auditor models judge whether an experiment was implemented faithfully and whether its evidence is valid. To measure how often they catch problems, we insert one known defect into a copy of a real experiment and see whether they notice. The copy is never used as a result.

The registered experiment:

<registration>
$registration
</registration>

The implementation, as submitted (code, scripts and any files it ships):

<files>
$files
</files>

The engineer's notes on the implementation:

<notes>
$notes
</notes>

The defect to insert:
$defect

Change the submitted files so that the copy has this defect when its evaluation command ($command) is run again on a fresh machine. Change only what the defect needs, keep the style of the code, and do not add comments that point to the change. The evaluation command must still run and write result.json. Edit only files shown in full above. Also rewrite the engineer's notes as they would read for the copy: the same honest-sounding account, with no mention of the defect and no result that the copy will not reproduce.

Reply with one JSON object in this form:

{"edits": [{"path": "the file's path as shown above", "content": "the whole new content of the file"}],
 "notes": "the engineer's notes for the copy",
 "description": "what you changed and where, for the study's records"}
\end{lstlisting}
\prompttitle{Review}{ (after the study)}
\begin{lstlisting}
Here is a research paper, as text taken from a web page. Its figures are not shown and its tables are broken into lines.

<seed_paper id="$arxiv_id">
$paper
</seed_paper>

You review one small experiment that was planned after this paper and then built and run. After the questions below you get its written plan (a "registration"), the notes of the engineer who built it, the output of its evaluation and the files the evaluation left.

The plan's writer had read this paper, could also read the paper's code and data and search the literature, and was told to choose their own question and method. The plan was written before anything was run. Every such plan had the same limits: one GPU for at most 45 minutes, and open models of 8B parameters or fewer. So it is a first small test. A separate engineer model then built the experiment from the paper and the registration, and the engineer's evaluation command was run once more on a fresh machine. You are not told which models wrote the plan or built the experiment.

You give eleven answers: seven about the idea, then four about what came out. The answers are used to count kinds of experiment, so choose the line that fits, not the one that sounds best. For each answer, write the reason in one sentence first, then the answer.

PART ONE: THE IDEA

The first seven answers are about the plan as it was written. Give them as you would have before the run: how the run came out does not make an idea more or less obvious, or its question more or less worth asking.

Four fields of the plan are the writer's own account of it: seed_connection, contribution, closest_work and full_scale_study. They say how the writer sees the link to the paper, what is new, and what larger study could follow. They are claims, not facts. Use them for one answer only, misreads_paper, where every statement the plan makes about the paper is checked against the paper. Give every other answer as you would if these four fields were empty.

These must not move any of the seven answers:
- how long, detailed or polished the text is, or how many controls it lists;
- whether you think the hypothesis is true, or how the run came out;
- that the experiment is small, or uses a smaller model or fewer examples than the paper because of the limits;
- whether this exact experiment has been published.

what_it_asks. One sentence of your own, taken from the hypothesis, the method and the event: what is compared with what, on what, to learn what. Leave out sample sizes, controls, statistics and thresholds.

in_the_paper. Look through the whole paper, appendix included.
- "reported": the paper reports this experiment, or a number or comparison that already answers its hypothesis by the same kind of measurement.
- "named": the paper does not report it, and says in its own words that this question is left undone: as future work, as a limitation, or as an open question.
- "neither": anything else. A sentence that only describes what the paper did, admits a weakness, explains a result without measuring it, or calls for more work in general, is not "named".
For "reported" or "named", copy into "quote" the passage of the paper that shows it: between 5 and 40 words in a row, exactly as printed, with nothing left out. Running prose and prompt listings count; tables and formulas do not. If you cannot point to such a passage, the answer is "neither" and "quote" is null.

relation. What is the hypothesis about? Decide by what the experiment tests and uses, not by its controls or side analyses, and not by whether the text mentions the paper.
The hypothesis is about the paper's own claims or results:
- "reruns": that a reported result comes out again.
- "checks": that a claim or result the paper states holds up, or fails, under a control, a baseline, a test for a confound or a stricter measurement.
- "varies": that a result of the paper holds, or fails, on another model, dataset or language, with the paper's method unchanged.
- "explains": why a reported result arises. Choose this only when the result is taken as real; if the hypothesis is that the result comes from a confound or an artifact, choose "checks".
The hypothesis is about something the paper does not claim or report:
- "builds on": the paper's method, data, benchmark or result is a working part of the experiment, and is changed, extended or put to a use the paper does not make of it. A test of a property or a side effect of the paper's method that the paper makes no claim about belongs here: the paper's claims stand whichever way it comes out.
- "new direction": the experiment uses none of these. The paper only prompted the question.
If two lines fit, choose the one that fits the hypothesis better.

obvious. Picture ten researchers who work in this paper's area and know its literature. Each has just read the paper and writes down the ten follow-up experiments that come to mind first. On how many of the ten lists would the idea of this experiment appear? The idea is the question and the main comparison in your what_it_asks sentence, without the test bed and the measurement. A list contains the idea if it has an experiment that asks the same question with the same main comparison. A list that has only the broader worry does not. Ask what they would write before seeing this plan, not whether the idea looks natural once read.
- "most": six or more lists.
- "some": three to five.
- "few": one or two.
- "none": no list.
These researchers know the standard techniques of the area. That makes an idea built on one of them available to them; still count how many would put it among their first ten.
An elaborate design, an invented test bed, or the lack of a published version does not make an idea less obvious. Being simple to run, or looking likely to work, does not make it more obvious.
A check of the paper can be an idea few would list. A step away from the paper can be one most would list.

who_would_care. Suppose the question behind this experiment, at the scope its hypothesis states, were followed through to a clear answer. Of the outcomes it could have, take the one that would change more. Who would then do or believe something different?
- "nobody".
- "a few groups": only people who work with the one method, benchmark or system the experiment is about.
- "one topic": people across a research topic would change how they do their own work, whatever method or benchmark they use.
- "several topics": people across several topics; a common practice or belief would change.
Count the people who would act on the answer. A question is not wide because it mentions a large field. How close the question is to the paper does not matter here.

misreads_paper. Does the plan depend on a statement about the paper (what it did, found or claimed) that the paper's text shows to be false? Look at every field, the writer's own account included.
- "yes": copy into "quote" the passage of the paper that shows it, by the same rule as above. If you cannot point to such a passage, the answer is "no" and "quote" is null.
- "no".
- "nothing stated": the plan states nothing about the paper that can be checked against its text.
Doubting or testing a claim of the paper is not a misreading. Repeating the paper's own summary of itself is not one either. The writer could see the paper's code and data and you cannot: a file name, a released model or a setting that the paper's text does not mention is not an error, and where the plan follows the code and the code differs from the text, that is not an error.

answerable. As it was designed, could this experiment answer the question in your what_it_asks sentence? Go by everything the plan says it will report, not only its one registered number.
- "yes": a clear result either way would bear on the question.
- "only if": the result can be read only if something else holds first, which the design cannot make sure of: for example that an effect of the paper appears at this small scale, that a model trained in the experiment learns what it is trained for, or that a measurement or a control behaves as it should. This is the answer whether or not the plan checks that thing itself. Name the one such thing that matters most in "needs", in a few words.
- "no": the design cannot answer the question as written, for example because the comparison leaves out what it claims to test, or no number it could produce could be read.
Judge the design as written, not by whether the run happened to work. Do not judge whether the hypothesis is likely to be true. Whether this many cases can give a clear result is not asked here: small size alone is not a reason for "no", and "the effect may be too small to see" is not a reason for "only if".

PART TWO: WHAT CAME OUT

The last four answers are about the run. A good plan can give a run that tells nothing, and a weak plan a run that tells something.

The engineer's notes are the engineer's own account: go by the numbers in the evaluation output and the files. Numbers the evaluation computed count, including checks the engineer added; where you can, check them against the raw outputs shown. Files marked "submitted, not written by the evaluation" came with the engineer's submission, for example logs and results of the engineer's own earlier runs. They are not results of the evaluation, and the engineer's notes may also quote those earlier runs.

The plan registered an event: one number set against one bar. The output says whether it occurred. None of the four answers is that. Writers set bars differently: a point value, a confidence bound, or a number written into the hypothesis. So a missed bar can come with clear evidence and a met bar with none. Look at how large the measured effect is and how sure the run is of it, and judge the same way whichever kind of bar the plan used.

Others have checked whether the code follows the plan. You need not list departures. Take the experiment as it was run.

what_came_out. One sentence with the numbers of the main comparison, how sure the run is of them (its interval if it gives one, otherwise the number of cases and repetitions), and the model and data. Do not say here whether the event occurred.

prerequisite. A test can rest on something that must hold before its main comparison can be read: that an effect of the paper appears at this scale, that a model trained in the experiment learned what it was trained for, or that a measurement or a control behaved as it should. It is something other than the hypothesis itself. If you named such a thing in "needs", look at it first; then look for any other.
- "held": as far as the run shows, nothing of this kind failed, or there was no such thing.
- "failed": one such thing did not hold, or did not pass the plan's own check for it, so the main comparison cannot be read. Name it in "which".
- "cannot tell": the run does not show whether such a thing held. Name it in "which".
With "failed" or "cannot tell", set "named_before" to true if "which" is what you named in "needs", otherwise false.
A fault that the run itself shows did not change the main comparison is not a failure. Too few cases or too much noise is not one either; that belongs under "evidence".

evidence. What does the run say about the plan's hypothesis? A number inside the hypothesis is the plan's bar. Do not answer by which side of it the result fell: a clear effect in the claimed direction, somewhat smaller than that number, is evidence for. If the hypothesis claims that something makes no difference or does not hold up, a run that clearly shows this is evidence for.
- "for": the run shows what the hypothesis claims, and noise alone would not explain it.
- "against": the run could have shown what the hypothesis claims and clearly shows otherwise. The change being tested was really applied and the measurement was not stuck at its floor or ceiling. For a claimed effect, the measured effect is near zero or reversed and an effect large enough to matter is ruled out.
- "mixed": the run is sure of two things that point opposite ways, for example an effect on one model and its reverse on another, or a gain with a measured loss.
- "inconclusive": the run does not tell either way: a prerequisite failed, or the uncertainty leaves room both for no effect and for an effect large enough to matter.
- "cannot judge": the numbers needed are not in what you are shown, or the code shows they do not measure what the plan describes.
If "prerequisite" is "failed", the answer is "inconclusive". If the hypothesis has two parts and one could not be tested, the answer is "inconclusive", not "mixed". If the run gives no interval, judge from the number of cases and repetitions and say so.

next_step. An experiment of this kind is a cheap first test of whether a question deserves a larger study. The question is the one in your what_it_asks sentence. Going only by what this run shows, what should a careful researcher do next about it?
- "larger study": the run justifies a study of many times its cost on the same question. It need not be the study in the plan's last field, and the evidence may be for or against the hypothesis.
- "one more experiment": the run answered its own question, and one gap must be closed by a further small experiment before choosing between a larger study and a stop. Name the experiment in "what".
- "repair": the run did not answer its own question and shows why. The same question must be asked again with the fault removed. Name the fault and the fix in "what". Too few cases is such a fault.
- "stop": the run justifies spending no more on this question, because (a) the idea was credibly shown not to work, or (b) something the whole question depends on was credibly shown not to hold, or (c) the question is settled and a larger study would add nothing that matters. Give the letter in "what".
- "no decision": the run did not answer its own question and does not show what to fix.
A justified stop is as good an answer as a justified larger study, and neither follows from the bar. If "evidence" is "inconclusive" or "cannot judge", the answer is "repair", "no decision", or "stop" on ground (b).

You have no tools: everything you can look at is in this message. Reply with one JSON object in this form and nothing else:

{"what_it_asks": "one sentence",
 "in_the_paper": {"reason": "one sentence", "answer": "reported | named | neither", "quote": null},
 "relation": {"reason": "one sentence", "answer": "reruns | checks | varies | explains | builds on | new direction"},
 "obvious": {"reason": "one sentence", "answer": "most | some | few | none"},
 "who_would_care": {"reason": "one sentence", "answer": "nobody | a few groups | one topic | several topics"},
 "misreads_paper": {"reason": "one sentence", "answer": "yes | no | nothing stated", "quote": null},
 "answerable": {"reason": "one sentence", "answer": "yes | only if | no", "needs": null},
 "what_came_out": "one sentence",
 "prerequisite": {"reason": "one sentence", "answer": "held | failed | cannot tell", "which": null, "named_before": null},
 "evidence": {"reason": "one sentence", "answer": "for | against | mixed | inconclusive | cannot judge"},
 "next_step": {"reason": "one sentence", "answer": "larger study | one more experiment | repair | stop | no decision", "what": null}}

The registration:

<registration>
$registration
</registration>

What was built and what came out:

<engineer_notes>
$notes
</engineer_notes>

<protected_evaluation>
$evaluation
</protected_evaluation>

<files>
$files
</files>
\end{lstlisting}

\end{document}